\documentclass[journal]{IEEEtran}

\PassOptionsToPackage{hidelinks}{hyperref}

\usepackage{cite}
\usepackage{amsmath,amssymb,amsfonts}
\usepackage{algorithmic}
\usepackage{graphicx}
\usepackage{textcomp}
\usepackage{xcolor}
\usepackage{orcidlink}
\usepackage{booktabs}
\usepackage{makecell}
\usepackage{multirow}
\usepackage{adjustbox}
\usepackage{enumitem}
\usepackage[switch]{lineno}
\usepackage{etoolbox}
\usepackage{subcaption}
\usepackage{hyperref}
\usepackage[normalem]{ulem}

\newcommand{\rev}[1]{#1}
\newcommand{\greenrev}[1]{#1}

\begin{document}
%\linenumbers
%\setlength\linenumbersep{8pt}
\title{From Physics to Surrogate Intelligence: A Unified Electro-Thermo-Optimization Framework for TSV Networks}

\author{%
Mohamed~Gharib$^*$\,\orcidlink{0009-0008-3606-6200},~\IEEEmembership{Graduate Student Member,~IEEE,~}%
Leonid~Popryho$^*$\,\orcidlink{0009-0002-0578-9592},~\IEEEmembership{Graduate Student Member,~IEEE,~}%
and~Inna~Partin-Vaisband\,\orcidlink{0000-0002-6399-6672},~\IEEEmembership{Senior Member,~IEEE}%
\thanks{$^*$These authors contributed equally to this work.}%
\thanks{Mohamed Gharib, Leonid Popryho, and Inna Partin-Vaisband are with the Department of Electrical and Computer Engineering, University of Illinois Chicago, Chicago, IL 60607, USA (e-mail: mghari5@uic.edu).}%
}

\markboth{}%
{Shell \MakeLowercase{\textit{et al.}}: Bare Demo of IEEEtran.cls for IEEE Journals}

\maketitle

\begin{abstract}
High-density through-substrate vias (TSVs) enable 2.5D/3D heterogeneous integration but introduce significant signal-integrity and thermal-reliability challenges due to electrical coupling, insertion loss, and self-heating. Conventional full-wave finite-element method (FEM) simulations provide high accuracy but become computationally prohibitive for large design-space exploration. 
This work presents a scalable electro--thermal modeling and optimization framework that combines physics-informed analytical modeling, graph neural network (GNN) surrogates, and full-wave sign-off validation. A multi-conductor analytical model computes broadband S-parameters and effective anisotropic thermal conductivities of TSV arrays, achieving $5\%$--$10\%$ relative Frobenius error (RFE) across array sizes up to $15\times15$. A physics-informed GNN surrogate (TSV-PhGNN), trained on analytical data and fine-tuned with HFSS simulations, generalizes to larger arrays \rev{with mean RFE below $5\%$ in-distribution}. 
The surrogate is integrated into a multi-objective Pareto optimization framework targeting reflection coefficient, insertion loss, worst-case crosstalk (NEXT/FEXT), and effective thermal conductivity. Millions of TSV configurations can be explored within minutes, enabling exhaustive layout and geometric optimization that would be infeasible using FEM alone. Final designs are validated with Ansys HFSS and Mechanical, showing strong agreement. The proposed framework enables rapid electro--thermal co-design of TSV arrays while reducing per-design evaluation time by more than six orders of magnitude.
\end{abstract}


\IEEEpeerreviewmaketitle

\section{Introduction}\label{sec:intro}

\greenrev{High-performance computing (HPC) and heterogeneous systems, including AI accelerators, memory-centric architectures, and sensor-integrated platforms, require high-bandwidth, low-loss, and thermally reliable interconnects for efficient die-to-die communication \cite{SALMA1, SALMA2, RAMI1}. While conventional 2D interconnects suffer from resistive/dielectric loss, bandwidth limits, and delay, through-substrate vias (TSVs) enable 3D heterogeneous integration through short, fine-pitch vertical paths that support high-bandwidth and low-latency communication \cite{9841607,11231028,10005194,11244450}. However, dense TSV arrays introduce reflection, insertion loss, and inter-TSV coupling \cite{6183642,doi:10.1142/8814,bookEDoTSV}, while dielectric liners reduce the substrate effective thermal conductivity (ETC), worsening thermal reliability and signal integrity \cite{6248927,7154441,9020021}. Thus, accurate early-stage electrothermal characterization is essential, although full-wave FEM tools such as Ansys HFSS are computationally prohibitive for large arrays, and mesh coarsening reduces fidelity.}

\begin{figure}[t]
\centering
\includegraphics[width=0.98\linewidth]{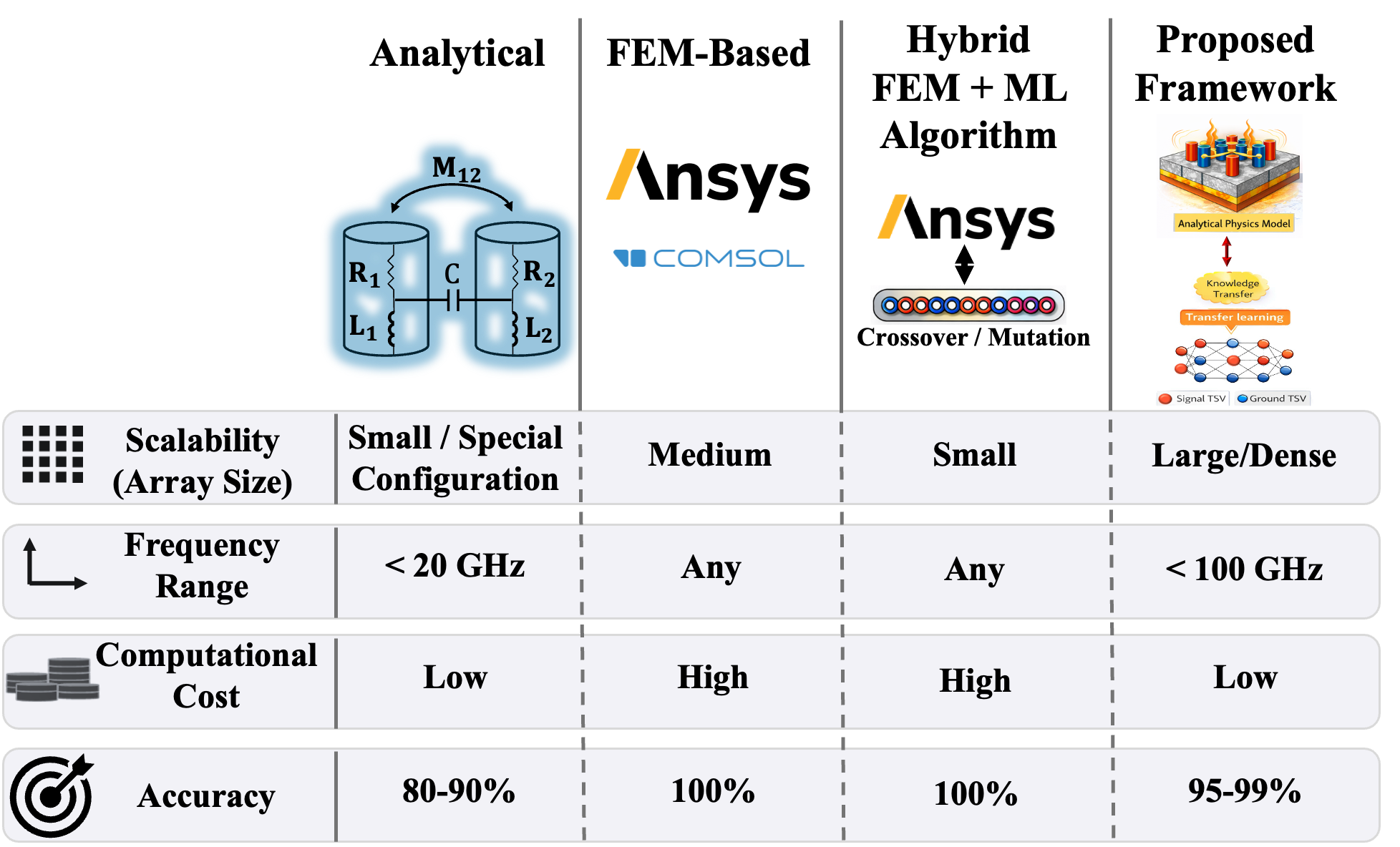}
\caption{Comparison of representative TSV modeling and optimization frameworks in terms of scalability, frequency range, and computational cost.}
\vspace{-10pt}
\label{fig:FIG1}
\end{figure}

\greenrev{Existing analytical and semi-analytical TSV models accelerate design-space exploration but are often limited to signal--ground pairs, small symmetric layouts, or simplified structures that cannot capture dense-array behavior \cite{10124931,10413605,10.1145/3649476.3658792,8031285,7480784,7434057}. Some also neglect parasitic inductance, restricting validity to below approximately 10--20~GHz \cite{7434057,7273940}. Practical TSV arrays are typically asymmetric signal--ground configurations with multiple return paths; although multi-ground structures were studied in \cite{7579236}, those models remain limited to symmetric layouts below 20~GHz and often assume ground rings around signal TSVs, neglecting inter-TSV coupling in dense arrays. ML-assisted FEM approaches have also been proposed: \cite{10830538} combines Ansys Q2D with genetic optimization for signal--ground assignment but is limited to $5\times5$ arrays, while tiled approximations for larger arrays ignore inter-sub-array coupling and can yield optimistic estimates, such as up to 13$\times$ lower coupling in a $10\times10$ array. Other multiphysics co-simulation, particle-swarm, and reinforcement-learning methods \cite{10198347,10403538} still rely on repeated FEM simulations and therefore scale poorly for dense TSV arrays \cite{10816184}. Thermal and electrothermal models reduce FEM dependence by treating TSVs as substrate conduction paths \cite{NIE2022114790,8382243,Sriharini_Minguin}, but many neglect high-frequency ohmic self-heating, which becomes significant in dense TSV structures. Fig.~\ref{fig:FIG1} summarizes representative TSV modeling and optimization frameworks in terms of scalability, frequency range, and computational cost.}

To address these limitations, this work presents a unified electro--thermal modeling and optimization framework for dense TSV arrays with arbitrary size, signal--ground allocation, and geometric parameters. The framework integrates physics-informed analytical modeling, graph-based machine learning (ML) acceleration, and full-wave FEM sign-off validation. The main contributions are summarized as follows:

\begin{itemize}[leftmargin=*]

\item \textbf{Physics-informed analytical electro--thermal modeling:}
A wideband analytical solver, extended from \cite{11038031} to support multi-ground TSV structures, computes multi-port S-parameters and effective anisotropic thermal conductivities ($k_x,k_y,k_z$) for arbitrary TSV layouts. Validation against HFSS using complex-domain relative Frobenius error (RFE) yields $5\%$--$10\%$ error while reducing simulation runtime from $\sim10^3$ seconds (FEM) to sub-second execution.

\item \textbf{Graph neural network surrogate modeling:}
A GNN-based surrogate trained on analytical data and fine-tuned with HFSS simulations generalizes from $3\times3$--$7\times7$ training arrays to grids up to $15\times15$, \rev{with mean RFE below $5\%$ in-distribution}. For a $15\times15$ array, the surrogate performs inference in $1.73~\mathrm{ms}$ compared with $2,998~\mathrm{s}$ for HFSS, corresponding to $\sim1.7\times10^6$ acceleration.

\item \textbf{Rapid large-scale design-space exploration:}
The surrogate enables exhaustive evaluation of large TSV configuration spaces. For a $5\times5$ grid with 12 signal TSVs, the $\binom{25}{12}=5.2\times10^6$ possible layouts can be explored in minutes. Exploiting dihedral symmetry ($D_4$) reduces the search space by up to $8\times$, enabling full exploration in under one minute. \rev{For larger arrays where exhaustive enumeration becomes intractable (e.g., $\binom{121}{58}\!\approx\!4.4\times10^{35}$ at $11\times11$), the same surrogate drives a multi-objective metaheuristic search such as NSGA-II at millisecond-per-candidate cost to construct Pareto fronts, as demonstrated in Sec.~\ref{results_and_sota}}.

\item \textbf{Multi-objective electro--thermal Pareto optimization:}
The framework jointly optimizes TSV layout and geometric parameters---including radius, pitch, height, and oxide thickness---to construct Pareto fronts capturing tradeoffs among reflection, insertion loss, crosstalk, and thermal conductivity.

\item \textbf{Automated FEM sign-off validation:}
Pareto-optimal designs are automatically exported through a PyAEDT workflow for verification using Ansys HFSS and Ansys Mechanical.

\end{itemize}

Overall, the proposed framework enables scalable electro--thermal co-design of dense TSV arrays, achieving million-fold acceleration over conventional FEM-based design-space exploration while maintaining near-FEM accuracy.

The remainder of this paper is organized as follows. Section~\ref{section_analyt} revisits the analytical framework of \cite{11038031} and extends it to multi-ground TSV structures. Section~\ref{ML_SECTION_EXPLAIN} describes the machine-learning methodology, including dataset generation, model architecture, and transfer learning strategy. Section~\ref{results_and_sota} evaluates the analytical and ML frameworks against full-wave FEM tools and state-of-the-art approaches. Finally, Section~\ref{conclusion} concludes the paper.

\section{Electrothermal Analytical Model}
\label{section_analyt}

% The electrothermal modeling framework for cylindrical vertical interconnects presented in~\cite{11038031} serves as the foundation of this work and is extended herein to support dense and irregular TSV arrays. In~\cite{11038031}, the electrical model is formulated under a metal--insulator--metal (MIM) coupling assumption and enables analytical extraction of self and mutual coupling terms for a single-reference ground configuration. In this work, the electrical formulation is generalized to multi-ground signal--ground arrangements, allowing accurate modeling of practical TSV arrays with arbitrary signal assignments and multiple return paths.

% On the thermal side,~\cite{11038031} employs a steady-state ETC approach that homogenizes the metal, liner, and substrate materials, with Joule heating modeled as volumetric heat sources. Here, the ETC formulation is modified to incorporate sparsity-aware corrections for partially populated and irregular TSV layouts. This anisotropic treatment preserves physical consistency and prevents systematic overestimation of lateral heat spreading that may arise from uniform homogenization assumptions.

The electrothermal modeling framework for cylindrical vertical interconnects in~\cite{11038031} is adopted as the basis of this work and extended to dense, irregular TSV arrays. Electrically, the original metal-insulator-metal (MIM) formulation for a single-reference ground is generalized to multi-ground signal-ground configurations, enabling accurate modeling of practical TSV arrays with arbitrary signal assignments and multiple return paths.

Thermally, the steady-state ETC formulation in~\cite{11038031} is extended with sparsity-aware corrections for partially populated and irregular layouts. This anisotropic treatment preserves physical consistency and avoids overestimating lateral heat spreading under uniform homogenization assumptions.

\subsection{Electromagnetic Framework}
\greenrev{The Python-based framework takes as input the substrate and TSV dimensions, operating frequency, and signal/ground assignments, and represents the array using the distributed RLCG equivalent-circuit formulation of~\cite{11038031}. That formulation provides closed-form expressions for the depletion thickness (obtained from the MOS-varactor sidewall-capacitance equation~\cite{wiley_book}), the oxide and depletion capacitances (coaxial transmission-line formulation~\cite{5613162}), the self and mutual external inductances (multiconductor transmission-line theory~\cite{paul2007analysis}), the substrate capacitance and conductance (obtained from the inductance formulation with permittivity and substrate conductivity replacing permeability), the frequency-dependent conductor impedance including skin-effect current crowding (via modified Bessel functions $I_0$ and $I_1$~\cite{paul2007analysis}), and the inter-metal-dielectric capacitance between adjacent TSVs (parallel-wire model~\cite{5739019}). The full equation set, the symbol glossary, and the SPICE-netlist generation flow are detailed in~\cite{11038031}; only the multi-ground extension that is new in this work is described below.}

In practical TSV arrays, multiple ground TSVs act as distributed return paths. To capture this behavior while reducing matrix dimensionality, the inductance matrix is constructed and reduced as follows:
\begin{enumerate}
    \item A single TSV is selected as a \textit{mathematical reference} for evaluating partial inductances $L_{ii}$ and $L_{ij}$. This choice only fixes the reference for inductance extraction and does not imply that current returns through a single TSV.
    
    \item The full inductance matrix $\mathbf{L}=\{L_{ij}\}$ is assembled and reordered into signal and ground sub-blocks,
    \begin{equation}
    \mathbf{L}=
    \begin{bmatrix}
    \mathbf{L}_{ss} & \mathbf{L}_{sg} \\
    \mathbf{L}_{gs} & \mathbf{L}_{gg}
    \end{bmatrix},
    \label{eq_1}
    \end{equation}
    where $\mathbf{L}_{ss}=\mathbf{L}[\mathcal{S},\mathcal{S}]$, 
    $\mathbf{L}_{sg}=\mathbf{L}[\mathcal{S},\mathcal{G}]$, 
    $\mathbf{L}_{gs}=\mathbf{L}[\mathcal{G},\mathcal{S}]$, and 
    $\mathbf{L}_{gg}=\mathbf{L}[\mathcal{G},\mathcal{G}]$, with $\mathcal{S}$ and $\mathcal{G}$ denoting the signal and ground index sets.

    \item The ground-current degrees of freedom are eliminated using a Schur-complement reduction \cite{paul2007analysis}, yielding the effective signal-domain inductance matrix
    \begin{equation}
        \mathbf{L}_{\mathrm{eff}} = \mathbf{L}_{ss} - \mathbf{L}_{sg}\mathbf{L}_{gg}^{-1}\mathbf{L}_{gs}.
        \label{eq_2}
    \end{equation}
    This operation captures current redistribution among the ground TSV network and preserves shielding and coupling effects while reducing the system dimension from $(N_s+N_g)$ to $N_s$.
\end{enumerate}

\greenrev{The same Schur-complement reduction is applied to the substrate capacitance and conductance matrices (obtained from the inductance formulation with $\epsilon_s$ and $\sigma_s$ replacing $\mu_0$), yielding effective signal-domain matrices that capture charge redistribution and shielding within the coupled ground TSV network. The complete RLCG netlist is then instantiated and simulated in PrimeSim HSPICE to extract the broadband S-parameters of the array.}

\subsection{Electrothermal Framework Using Equivalent Thermal Conductivity}
\label{electrothermal_model}

\begin{figure}[t]
  \centering
  \includegraphics[width=\linewidth]{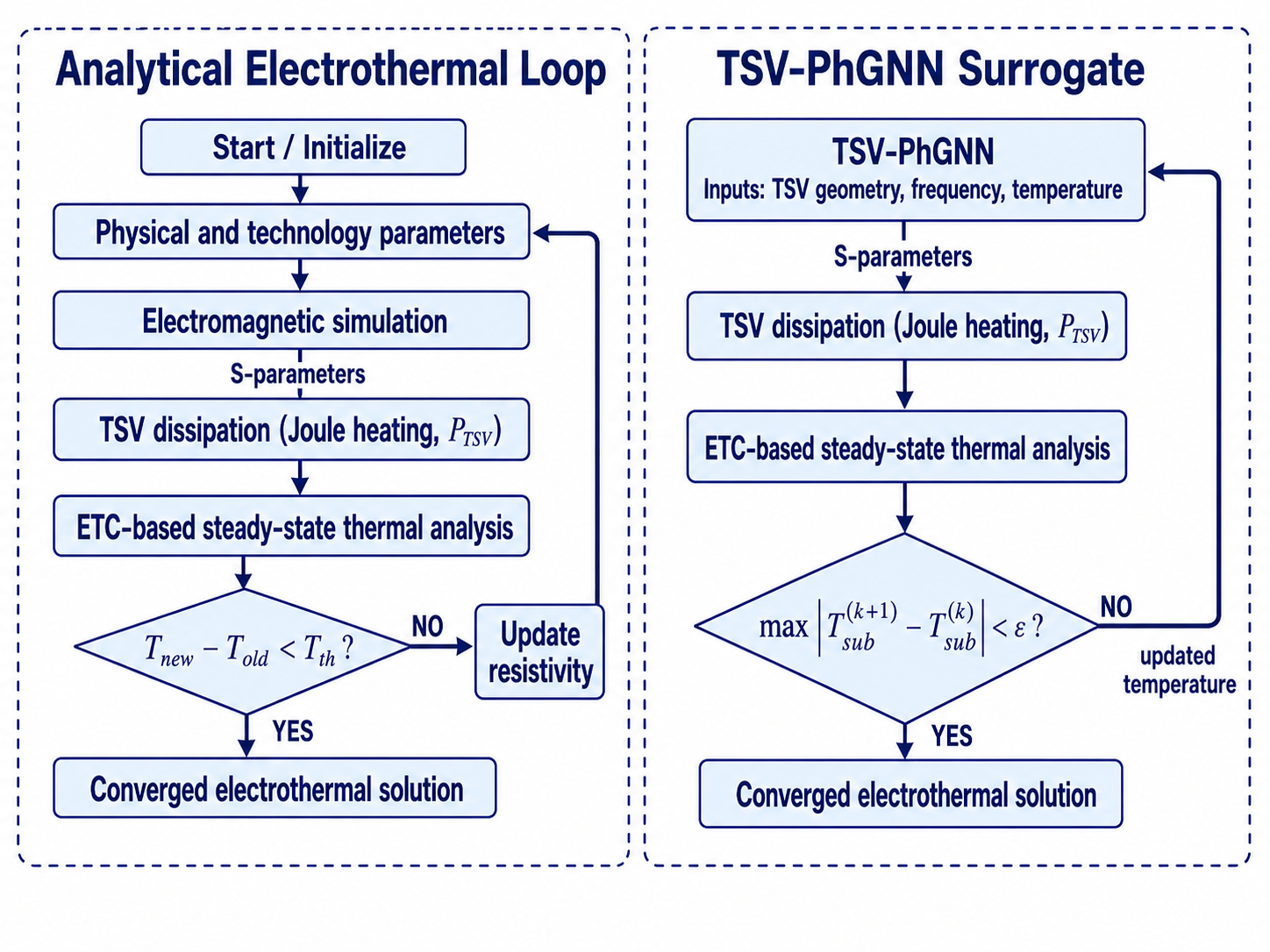}
  \vspace{-15pt}
  \caption{\rev{Comparison of the analytical and TSV-PhGNN-based coupled electrothermal simulation frameworks. In both cases, S-parameters are used to extract TSV Joule-heating dissipation, which is then mapped to the ETC-based steady-state thermal solver. In the analytical loop, temperature-dependent resistivity is updated iteratively, whereas in the surrogate loop, the updated temperature is fed back to TSV-PhGNN, until convergence is reached.}}
  \vspace{-5pt}
  \label{fig:Flow}
\end{figure}

\greenrev{Joule heating in the TSVs raises copper resistivity, which further amplifies the ohmic losses---a positive feedback loop that must converge to a steady-state operating point. Following the coupled electrothermal flow of~\cite{11038031} (reproduced in Fig.~\ref{fig:Flow}), electrical losses extracted from the S-parameters are mapped to volumetric heat sources, and the resulting temperature is used to update the temperature-dependent copper resistivity ($\rho_{\mathrm{Cu}}=\rho_0[1+\alpha(T-T_0)]$~\cite{Matula1979ElectricalRO}) in the next electrical solve. The focus of this work is on the steady-state operating point.}

\subsubsection{Equivalent Thermal Conductivity Formulation}

\greenrev{Each TSV is modeled as a square copper--oxide--substrate unit cell. The vertical and lateral per-cell equivalent thermal conductivities, $k^{z}_{\text{cond}}$ and $k^{x,y}_{\text{cond}}$, are computed using the area-weighted and integral formulations of~\cite{11038031}, which preserve nonuniform lateral heat-flow paths without resorting to simple series/parallel approximations. These per-cell values serve as the building blocks for the array-level homogenization described next.}

\subsubsection{Equivalent Volumetric Heat Capacity}

In addition to ETC, the framework computes an equivalent volumetric heat
capacity for completeness and future use in transient electrothermal
simulations. For a single TSV unit cell, the equivalent volumetric heat
capacity is given by
\begin{equation}
    (\rho C_p)_{\text{eq}} =
    \frac{
    \rho_v C_{p,v} s_v^z +
    \rho_l C_{p,l} s_l^z +
    \rho_s C_{p,s} s_s^z}
    {4(r_{\text{cond}} + t_{\text{ins}})^2},
\end{equation}
where $\rho$ and $C_p$ denote, respectively, the mass density and specific heat capacity of the corresponding material region, \greenrev{and $s_v^z$, $s_l^z$, $s_s^z$ are the per-cell cross-sectional areas of the via, liner, and substrate as defined in~\cite{11038031}}.

To account for the surrounding substrate beyond the TSV unit cells,
an extended equivalent volumetric heat capacity is computed over the full
substrate area. Under steady-state conditions, the $\rho C_p$ term
multiplies the transient temperature derivative and therefore drops out
from the governing heat equation. Nevertheless, it is reported to enable
direct extension of the framework to transient electrothermal analysis.

\subsubsection{Array-Level Anisotropic Conductivities}

\rev{
Prior array-level equivalent thermal conductivity (ETC) formulations,
including \cite{NIE2022114790} and \cite{11038031}, assume a fully populated rectangular TSV array with uniform spacing, such that all $M \times N$ unit cells contain TSVs. While this assumption simplifies homogenization, practical TSV layouts are often sparse and irregular due to routing constraints, power-delivery structures, and mechanical keep-out regions. Consequently, direct application of fully populated ETC models can overestimate the lateral heat-spreading capability.
}

\rev{
To address this limitation, the proposed framework extends the
array-level ETC formulations in \cite{NIE2022114790} to sparse TSV
arrays through an occupancy-aware homogenization approach.
Let $M$ and $N$ denote the number of rows and columns of the bounding TSV grid, respectively, and let $n_{\text{TSV}}$ denote the total number of existing TSVs. A TSV occupancy factor is defined as
}
\begin{equation}
    f_{\text{occ}} = \frac{n_{\text{TSV}}}{MN},
    \qquad 0 < f_{\text{occ}} \le 1 .
\end{equation}

\rev{
To preserve the correct total TSV area fraction during lateral thermal
homogenization, the effective TSV footprint is scaled by
$\sqrt{f_{\text{occ}}}$. To the best of our knowledge, this $\sqrt{f_{\text{occ}}}$ occupancy-aware correction is introduced in this work; prior array-level ETC formulations~\cite{NIE2022114790,11038031} are derived for fully populated arrays and do not include an explicit sparsity factor. The scaling yields
}

\begin{equation}
    l_{\text{cond,eff}} = w_{\text{cond,eff}}
    =
    2(r_{\text{cond}} + t_{\text{ins}})
    \sqrt{f_{\text{occ}}},
\end{equation}

\rev{
where $r_{\text{cond}}$ and $t_{\text{ins}}$ denote the TSV conductor
radius and insulation thickness, respectively. The square-root scaling
ensures that the effective conductive area scales proportionally to
$f_{\text{occ}}$.
Using these effective dimensions, the lateral ETCs are expressed as
}
\rev{
{\footnotesize
\begin{equation}
\label{eq.12}
\thinmuskip=0mu
\medmuskip=0mu
\thickmuskip=0mu
k^{x}_{eq} =
\frac{k_{s} (w_s - Nw_\text{cond,eff})}{w_s}
+ \left(
\frac{k_s Nl_sw_\text{cond,eff}}
{(l_s - Ml_\text{cond,eff})w_s}
\bigg| \bigg|
\frac{k^{x}_\text{cond} Nl_sw_\text{cond,eff}}
{Ml_\text{cond,eff}w_s}
\right),
\end{equation}
}

{\footnotesize
\begin{equation}
\label{eq.13}
\thinmuskip=0mu
\medmuskip=0mu
\thickmuskip=0mu
k^{y}_{eq} =
\frac{k_{s} (l_s - Ml_\text{cond,eff})}{l_s}
+ \left(
\frac{k_s Ml_\text{cond,eff}w_s}
{(w_s - Nw_\text{cond,eff})l_s}
\bigg| \bigg|
\frac{k^{y}_\text{cond} Ml_\text{cond,eff}w_s}
{Nl_sw_\text{cond,eff}}
\right),
\end{equation}
}
}

\rev{
\noindent where $\|$ denotes parallel thermal conduction paths and
$l_s$ and $w_s$ denote the substrate lateral dimensions.
The first terms model heat conduction through the silicon region,
while the second term represents equivalent parallel heat flow through
silicon and TSV conductive paths.
}
\rev{
The vertical ETC is computed using the actual TSV area fraction rather
than assuming a fully populated array,
}
\begin{equation}
\label{eq.14}
    k^{z}_{eq}
    =
    k_s
    +
    \frac{l_{\text{cond}}w_{\text{cond}}}
    {l_s w_s}
    n_{\text{TSV}}
    \left(k^{z}_{\text{cond}}-k_s\right).
\end{equation}

\rev{
Unlike prior formulations that implicitly assume uniform TSV occupancy, the proposed sparsity-aware ETC model preserves the physical TSV volume fraction and avoids systematic overestimation of thermal conductivity in partially populated and irregular TSV layouts.
}

\subsubsection{Heat Source Extraction from S-Parameters}

\greenrev{The volumetric heat-generation field $G$ is derived from the S-parameters as in~\cite{11038031}: for each excited port the dissipated power equals the deficit between input and outgoing scattered power,}
\begin{equation}
    P_{\text{loss}} = P_{\text{in}}
    \left(1 - \sum_{i=1}^{q} |S_{ij}|^2 \right),
    \label{pinminuspout}
\end{equation}
\greenrev{with $q=(N\times M-1)\times 2$ ports in total (each TSV modeled as a two-port). Per-conductor $P_{\text{loss}}$ values are normalized over the corresponding TSV volume and assembled into the volumetric heat-generation field $G$ used as the source term for the thermal block.}

\subsubsection{Steady-State Thermal Solution Using Ansys Mechanical}

% Rather than explicitly solving the heat conduction equation on a full TSV-resolved geometry, the derived anisotropic ETCs $(k^{x}_{eq}, k^{y}_{eq}, k^{z}_{eq})$ are imported into \emph{Ansys Mechanical} by defining an equivalent anisotropic thermal block. The volumetric heat generation field computed from S-parameters is applied directly to this block, and identical boundary conditions (Dirichlet or convective) to those used in full 3D simulations are enforced.

% This homogenized ETC-based approach enables rapid steady-state temperature extraction while maintaining close agreement with explicit TSV-resolved finite-element thermal simulations, achieving a computational speedup ranging from approximately $35\times$ for $5\times5$ sparse arrays to about $1300\times$ for fully populated $10\times10$ arrays  compared to full 3D thermal analysis.
Instead of explicitly solving the heat conduction equation on TSV-resolved geometries, the extracted anisotropic ETCs $(k^{x}_{eq}, k^{y}_{eq}, k^{z}_{eq})$ are imported into \emph{Ansys Mechanical} as an equivalent anisotropic thermal block. The volumetric heat sources derived from the S-parameters are applied to this block under the same boundary conditions used in full 3D simulations.

This homogenized ETC-based approach enables rapid steady-state temperature extraction while maintaining close agreement with TSV-resolved finite-element simulations, achieving speedups from $\sim35\times$ for $5\times5$ sparse arrays to $\sim1300\times$ for fully populated $10\times10$ arrays.
\section{ML-Assisted Optimization Framework}
\label{ML_SECTION_EXPLAIN}

The proposed framework integrates analytical physics modeling, GNNs, and transfer learning to create a surrogate model capable of rapid and accurate S-parameter prediction for TSV arrays.

\subsection{Analytical Dataset Generation}
To mitigate the prohibitive computational cost of three-dimensional FEM simulations (e.g., Ansys HFSS) during the initial model training stage, the physics-based analytical framework described in Section~\ref{section_analyt} is used to generate a large-scale pretraining dataset. Using this framework, 100,000 data samples are generated analytically for TSV arrays with dimensions ranging from $3\times 3$ to $ 20\times 20$, and the corresponding RLCG-based netlists are simulated using the HSPICE circuit simulator. The resulting dataset is partitioned into 80,000 samples for training and 20,000 samples for validation.

\subsection{Graph Representation}

The TSV array is modeled as a graph $\mathcal{G} = (\mathcal{V}, \mathcal{E})$, enabling a flexible representation that naturally supports arbitrary TSV arrangement patterns and variable array sizes. Graph-based modeling is particularly well suited for this problem, as it captures complex spatial interactions while remaining inherently scalable. As a result, the proposed framework can generalize to TSV arrays larger than those encountered during training without requiring architectural modifications.

\subsubsection{Node Representation}

\rev{The node set $\mathcal{V}$ corresponds to the TSVs in the array, where each TSV is represented by a node $v_i \in \mathcal{V}$. The node feature vector $\mathbf{x}_i \in \mathbb{R}^7$ encodes the TSV electrical role, geometric parameters, operating conditions, and thermal state:
\begin{equation}
    \mathbf{x}_i = \big[\text{Type},\, r_{\text{via}},\, p,\, h_{\text{via}},\, t_{\text{ox}},\, f,\, Temp \big].
\end{equation}
Here, $\text{Type} \in {1,0,-1}$ identifies signal TSVs, empty locations, and ground TSVs, respectively. The remaining features correspond to the TSV radius $r_{\text{via}}$, pitch $p$, via height $h_{\text{via}}$, oxide thickness $t_{\text{ox}}$, operating frequency $f$, and TSV temperature $Temp$. By jointly incorporating structural, electrical, and thermal information, the node representation enables the GNN to capture the coupled electrothermal behavior of TSV arrays and learn interactions that vary with geometry, excitation conditions, and local temperature.}

\subsubsection{Edge Representation}

The edge set $\mathcal{E}$ captures electromagnetic interactions between TSVs. To account for long-range coupling effects, the graph is constructed as fully connected, where each edge $e_{ij} \in \mathcal{E}$ represents the interaction between TSVs $i$ and $j$. The corresponding edge feature vector $\mathbf{e}_{ij}$ encodes their spatial relationship through distance-based descriptors:
\begin{equation}
    \mathbf{e}_{ij} = \big[d_{ij},\, d_{ij}^{-1},\, d_{ij}^{-2}\big],
\end{equation}
where $d_{ij}$ is the Euclidean distance between TSVs $i$ and $j$. Explicit inclusion of inverse-distance terms facilitates learning of electromagnetic field decay behavior and mutual coupling characteristics, which are critical for accurate wideband S-parameter prediction.

% \begin{figure}[htbp]
%     \centering
%     \includegraphics[width=0.95\linewidth]{images/Graph_w_temp.png}
%     \caption{\rev{Graph representation of a $3\times4$ TSV array. Red nodes represent Signal TSVs, and blue nodes represent Ground TSVs. Edges carry spatial distance features.}}
%     \label{fig:graph_rep}
% \end{figure}

\subsection{Physics-Informed Graph Neural Network}

The proposed learning architecture, referred to as \emph{TSV-PhGNN}, is explicitly designed to embed key physical principles governing electromagnetic propagation and coupling in TSV arrays.

\subsubsection{Feature-Wise Linear Modulation}

\rev{TSV interconnects operate across distinct electromagnetic regimes whose boundaries are set jointly by frequency and temperature. At high frequency, current crowds into a skin layer of thickness $\delta(f,T)=1/\sqrt{\pi f \mu \sigma(T)}=\sqrt{\rho(T)/(\pi f \mu)}$, where the metallic resistivity follows the linear law $\rho(T)=\rho_0[1+\alpha(T-T_{\mathrm{ref}})]$. Substituting $\delta$ into the per-unit-length AC resistance $R_{\mathrm{ac}}\propto \rho(T)/\delta$ yields the leading-order scaling
\begin{equation}
    R_{\mathrm{ac}}(f,T)\;\propto\;\sqrt{f\,\rho(T)}\;\approx\;\sqrt{f\bigl[1+\alpha(T-T_{\mathrm{ref}})\bigr]},
\end{equation}
which couples $f$ and $T$ multiplicatively. Treating $f$ and $T$ as ordinary input features would force the network to learn this cross-coupling from scratch; they are instead promoted to \emph{conditioning} variables, and the product is injected explicitly through Feature-wise Linear Modulation (FiLM)~\cite{10.5555/3504035.3504518}.
The conditioning vector is
\begin{equation}
    \mathbf{c}=[\,f,\;T,\;f\!\cdot\!T\,],\qquad
    \mathbf{h}_{\text{cond}}=\mathrm{MLP}(\mathbf{c}),
\end{equation}
and the geometry embedding $\mathbf{h}_{\text{geom}}$ is modulated as
\begin{equation}
    \mathbf{h}'=\gamma(\mathbf{h}_{\text{cond}})\odot\mathbf{h}_{\text{geom}}+\beta(\mathbf{h}_{\text{cond}}),
\end{equation}
with $\gamma(\cdot),\beta(\cdot)$ learnable affine maps and $\odot$ denoting the Hadamard product. Because FiLM is linear in $\mathbf{h}_{\text{cond}}$, embedding the $f\!\cdot\!T$ term directly supplies the inductive bias needed to reproduce the $\sqrt{f\rho(T)}$ loss trend and to extrapolate along the physically correct direction when the closed electro-thermal loop drives the operating temperature beyond the densely sampled training regime. }

\subsubsection{Message Passing via Graph Transformer}

Message passing is performed using a Graph Transformer architecture. Unlike uniform aggregation schemes, the Transformer computes attention coefficients $\alpha_{ij}$ for each edge, allowing the model to learn direction-dependent and context-aware coupling strengths between TSVs. This mechanism is particularly effective for capturing electromagnetic shielding and screening effects, for example by assigning reduced attention weights to interactions obstructed by ground TSVs.

\subsubsection{Symmetry-Enforced Output Heads}

The network predicts complex-valued S-parameters and employs two specialized output heads:
\begin{itemize}
    \item \textbf{Node Head:} Predicts return loss ($S_{11}$) and insertion loss ($S_{21}$) for each signal TSV.
    \item \textbf{Edge Head:} Predicts near-end crosstalk (NEXT) and far-end crosstalk (FEXT) between pairs of signal TSVs.
\end{itemize}

A fundamental physical constraint is reciprocity, which requires $S_{ij} = S_{ji}$. Since standard neural network architectures do not inherently satisfy this condition, reciprocity is enforced explicitly by symmetrizing the edge-level predictions:
\begin{equation}
    \hat{y}_{\text{edge}}^{(i,j)} =
    \frac{1}{2}
    \left(
    \text{MLP}(\mathbf{h}_i, \mathbf{h}_j)
    +
    \text{MLP}(\mathbf{h}_j, \mathbf{h}_i)
    \right).
\end{equation}

\begin{figure*}[htbp]
    \centering
    \includegraphics[width=0.95\linewidth]{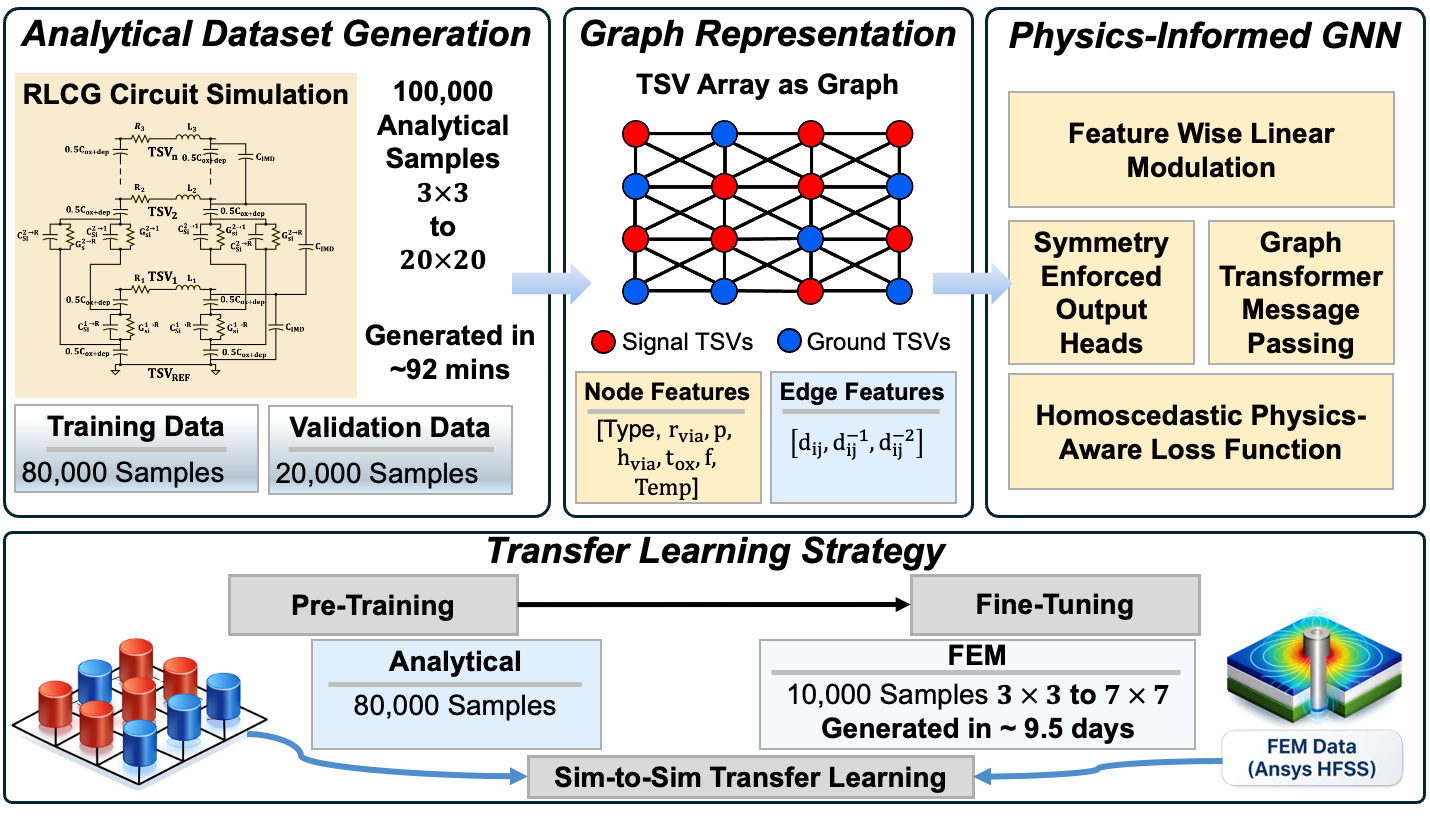}
\caption{\rev{Overview of the proposed ML-assisted optimization framework. A physics-informed GNN is first pre-trained on large-scale RLCG-based analytical data and then fine-tuned on high-fidelity Ansys HFSS data through Sim-to-Sim transfer learning. The TSV array is modeled as a geometry-, frequency-, and temperature-aware graph, and the trained surrogate is embedded in the coupled electrothermal flow shown in Fig.~\ref{fig:Flow}.}}

    \label{fig:MODEL_ML}
\end{figure*}

\vspace{-10pt}
\subsection{Homoscedastic Physics-Aware Loss Function}

The training objective jointly enforces regression accuracy and physical consistency. Due to the large dynamic range between transmission-related terms (e.g., $S_{21}$) and crosstalk terms (e.g., $S_{ij}$), direct optimization using a single loss leads to task imbalance. To address this issue, a homoscedastic uncertainty-based weighting strategy is adopted.
The regression loss is defined as
\begin{equation}
\mathcal{L}_{\text{reg}} =
\sum_{k \in \mathcal{K}}
\left(
\frac{1}{2\sigma_k^2}\,\mathcal{L}_{\text{MSE},k}
+ \log \sigma_k
\right),
\end{equation}
where $\mathcal{K}=\{\mathrm{IL},\mathrm{RL},\mathrm{NEXT},\mathrm{FEXT}\}$ denotes the set of predicted S-parameter quantities. Specifically, $\mathrm{IL}$ corresponds to insertion loss (e.g., $|S_{21}|$ in dB), $\mathrm{RL}$ to return loss (e.g., $|S_{11}|$ in dB), and $\mathrm{NEXT}$/$\mathrm{FEXT}$ to near- and far-end crosstalk magnitudes (e.g., $|S_{31}|$ and $|S_{41}|$ in dB, depending on port assignment).

The mean-squared error for task $k$ is defined as
\begin{equation}
\mathcal{L}_{\mathrm{MSE},k}
=
\frac{1}{N}
\sum_{n=1}^{N}
\left(
\hat{y}_{k,n} - y_{k,n}
\right)^2,
\end{equation}
where $\hat{y}_{k,n}$ and $y_{k,n}$ denote the predicted and reference values, respectively. The parameters $\sigma_k$ are learnable task-dependent variances that adaptively balance the contributions of the individual regression targets during training.

% The regression loss is defined as
% \begin{equation}
%     \mathcal{L}_{\text{reg}} =
%     \sum_{k \in \{\text{Trans}, \text{Ref}, \text{NEXT}, \text{FEXT}\}}
%     \left(
%     \frac{1}{2\sigma_k^2} \mathcal{L}_{\text{MSE},k}
%     + \log \sigma_k
%     \right),
% \end{equation}
% where $\mathcal{L}_{\mathrm{MSE},k} = \frac{1}{N}\sum_{n=1}^{N} \left( \hat{y}_{k,n} - y_{k,n} \right)^2$
% is the mean squared error for task $k$, and $\sigma_k$ are learnable variance parameters that automatically balance the contributions of the individual tasks.

\rev{In addition, a \emph{passivity constraint} is incorporated into the training objective. For a passive linear multiport network, the scattering matrix must satisfy $\sigma_{\max}(\mathbf{S}) \leq 1$, where $\sigma_{\max}$ denotes the largest singular value of $\mathbf{S}$~\cite{anderson1973}. Accordingly, the passivity loss is defined as
\begin{equation}
\mathcal{L}_{\text{passivity}}
=
\operatorname{mean}_{s}
\operatorname{ReLU}
\bigl(
\sigma_{\max}(\mathbf{S}^{(s)})-1
\bigr),
\end{equation}
where $\mathbf{S}^{(s)}$ is assembled for each sample from the predicted node ($S_{11},S_{21}$) and edge ($S^{\mathrm{NEXT}},S^{\mathrm{FEXT}}$) outputs. This differentiable spectral penalty directly enforces the physical passivity condition by penalizing only singular values that exceed unity. The total training objective is given by
\begin{equation}
\mathcal{L}
=
\mathcal{L}_{\text{reg}}
+
\lambda
\mathcal{L}_{\text{passivity}},
\end{equation}
where $\lambda$ controls the strength of the passivity regularization.}

\subsection{Transfer Learning Strategy}

While the implemented physics-based analytical model enables rapid generation of large-scale training data, it relies on simplifying assumptions (e.g., quasi-TEM behavior) that do not fully capture three-dimensional electromagnetic effects. To achieve analytical scalability with full-wave fidelity, a multi-fidelity simulation-to-simulation (Sim-to-Sim) transfer learning framework is adopted~\cite{transfer_learning}. The approach exploits the strong correlation between low-fidelity (analytical) and high-fidelity (full-wave) physical manifolds. Although the analytical model may exhibit magnitude deviations, it preserves the dominant geometric and frequency-dependent trends of the S-parameters. Accordingly, training is performed in two stages.

\rev{\textit{1) Pre-training (Low Fidelity):} The proposed TSV-PhGNN is first trained on a large analytically generated dataset (${N=80{,}000}$) obtained from RLCG-based HSPICE simulations. The dataset spans operating temperatures from 300 K to 600 K, enabling the network to learn topology-aware electrothermal relationships between TSV geometry, spatial arrangement, operating conditions, and broadband S-parameter behavior. This stage provides a physics-consistent initialization for subsequent transfer learning.}

\rev{\textit{2) Fine-tuning (High Fidelity):} The pre-trained model is subsequently fine-tuned using a smaller full-wave dataset ($N=10{,}000$) generated from 3D FEM simulations in Ansys HFSS for compact arrays ($3\times3$ to $7\times7$) over the same 300 K--600 K temperature range. Using an identical input feature space and a reduced learning rate ($10^{-4}$), the network adapts to higher-order electromagnetic and electrothermal effects while retaining the structural knowledge acquired during pre-training.}

\rev{In both stages, each sample is generated at a single prescribed $T$, with $\rho(T)=\rho_{0}[1+\alpha(T-T_{\mathrm{ref}})]$ applied per sample to the HSPICE netlist or HFSS material model. The surrogate is therefore trained on a $(\text{geometry}, f, T)$ snapshot family rather than on per-sample converged electrothermal fixed points; the coupled-loop fixed point is reconstructed at inference via the iteration of Section~\ref{Unified_section}.}

This multi-fidelity strategy substantially reduces reliance on computationally intensive full-wave simulations while maintaining FEM-level predictive accuracy (Fig.~\ref{fig:MODEL_ML}). Thermal parameters, including the equivalent thermal conductivities, are computed directly from the analytical electrothermal model (Section~\ref{electrothermal_model}) since their evaluation is computationally inexpensive and does not require surrogate modeling.

\subsection{Unified Electro-Thermal Framework}
\label{Unified_section}

\rev{The proposed framework iteratively couples the TSV-PhGNN electrical
surrogate with the ETC-based thermal solver. At iteration $k$, the
TSV-PhGNN uses the TSV graph, operating frequency $f$, and temperature
feature $T^{(k)}$ to predict the array scattering matrix
$\widehat{\mathbf{S}}^{(k)}$. The absorbed TSV Joule power
$P_{\mathrm{TSV}}^{(k)}$ is then computed from the predicted
S-parameters using Eq.~(\ref{pinminuspout}).}

\rev{The extracted power $P_{\mathrm{TSV}}^{(k)}$ is applied as a
heat-generation source in the homogenized TSV-substrate region of the
thermal solver, together with the prescribed boundary conditions.}

\rev{The ETC-based solver computes the updated substrate-averaged
temperature $\overline{T}_{\mathrm{sub}}^{(k+1)}$ using the anisotropic
array-level ETC formulation $\bigl(k^{x}_{eq},k^{y}_{eq},k^{z}_{eq}\bigr)$
in Eqs.~(\ref{eq.12})--(\ref{eq.14}). Since individual TSVs are not
thermally resolved, this averaged temperature is assigned to all TSV nodes
as the next temperature feature $T^{(k+1)}$.}

\rev{Accordingly, the electrical surrogate updates the thermal problem
through the S-parameter-derived dissipation, while the thermal solver
updates the electrical problem through temperature feedback. As shown in
Fig.~\ref{fig:Flow}, the iteration continues until}
\begin{equation}
\rev{
\left|
\overline{T}_{\mathrm{sub}}^{(k+1)}
-
\overline{T}_{\mathrm{sub}}^{(k)}
\right|
\leq
0.5~\mathrm{K}.
}
\end{equation}

\section{Experiments and Results}
\label{results_and_sota}

This section presents a comprehensive evaluation of the proposed framework. First, the analytical electrothermal model described in Section~\ref{section_analyt} is validated against full-wave electromagnetic simulations in Ansys HFSS and thermal simulations in Ansys Mechanical. This comparison establishes the accuracy and reliability of the analytical framework, thereby justifying its use for large-scale dataset generation during the pre-training stage of the ML model prior to fine-tuning with high-fidelity Ansys data.
Second, the predictive performance of the proposed ML model is evaluated by comparing its S-parameter predictions against Ansys HFSS simulation results. \rev{In addition, a fully coupled electrothermal testcase is validated against a bidirectional Ansys HFSS--Mechanical co-simulation to assess the convergence behavior and accuracy of the proposed coupled framework.}
Finally, optimization case studies are conducted using the trained ML surrogate model, and the resulting designs are benchmarked against SOTA TSV architectures to demonstrate performance improvements.

\subsection{Experimental Setup}
All circuit-level simulations are performed using Synopsys PrimeSim HSPICE (V-2023.12-2, Linux64) on a high-performance compute server running Red Hat Enterprise Linux 9.7 (Plow). The server is equipped with an AMD Ryzen 9 7950X processor (16 physical cores, 32 threads), 128~GB of system memory, and an NVIDIA GeForce RTX 4090 GPU with 24~GB of VRAM. Machine learning training and inference are conducted in Python~3.13.5 using PyTorch~2.8.0 with CUDA~12.9 support.
Full-wave electromagnetic and thermal simulations are carried out using Ansys Electronic Desktop (AEDT) 2024~R2 on a separate workstation running Windows~11. 
The system features a 14-core (20-thread) Intel\textsuperscript{\textregistered} processor operating at up to 2.6~GHz and 32~GB of memory.
%This system is equipped with an Intel\textsuperscript{\textregistered} processor featuring 14 physical cores (20 logical threads) operating at a maximum clock frequency of 2.6~GHz, and 32~GB of system memory. 
All HFSS and Mechanical simulations are executed within a Python~3.12 virtual environment.

To ensure a fair runtime comparison between the proposed ETC-based framework and the full EM--thermal simulation flow, both approaches are executed under identical hardware and software configurations on their respective platforms.

\subsection{Analytical Model Performance}

\begin{figure}[t]
    \centering
    \includegraphics[width=0.95\linewidth]{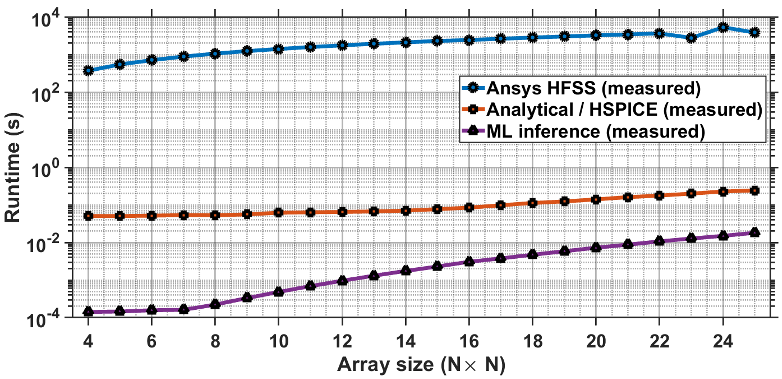}
    \caption{\rev{Per-sample runtime comparison between the ML surrogate, the analytical solver, and Ansys HFSS versus TSV grid size ($N\times N$). }}
    \label{fig:Inf_time}
\end{figure}

The proposed electromagnetic analytical framework is evaluated on TSV arrays ranging from $4\times4$ to $25\times25$ using 1,000 randomly generated samples. 
As shown in Fig. \ref{fig:Inf_time}, the average runtime increased modestly from 27~ms for the $4\times4$ array to 244~ms for the $25\times25$ array, demonstrating favorable scalability with array size. 

\begin{figure}[t]
    \centering
    \includegraphics[width=\linewidth]{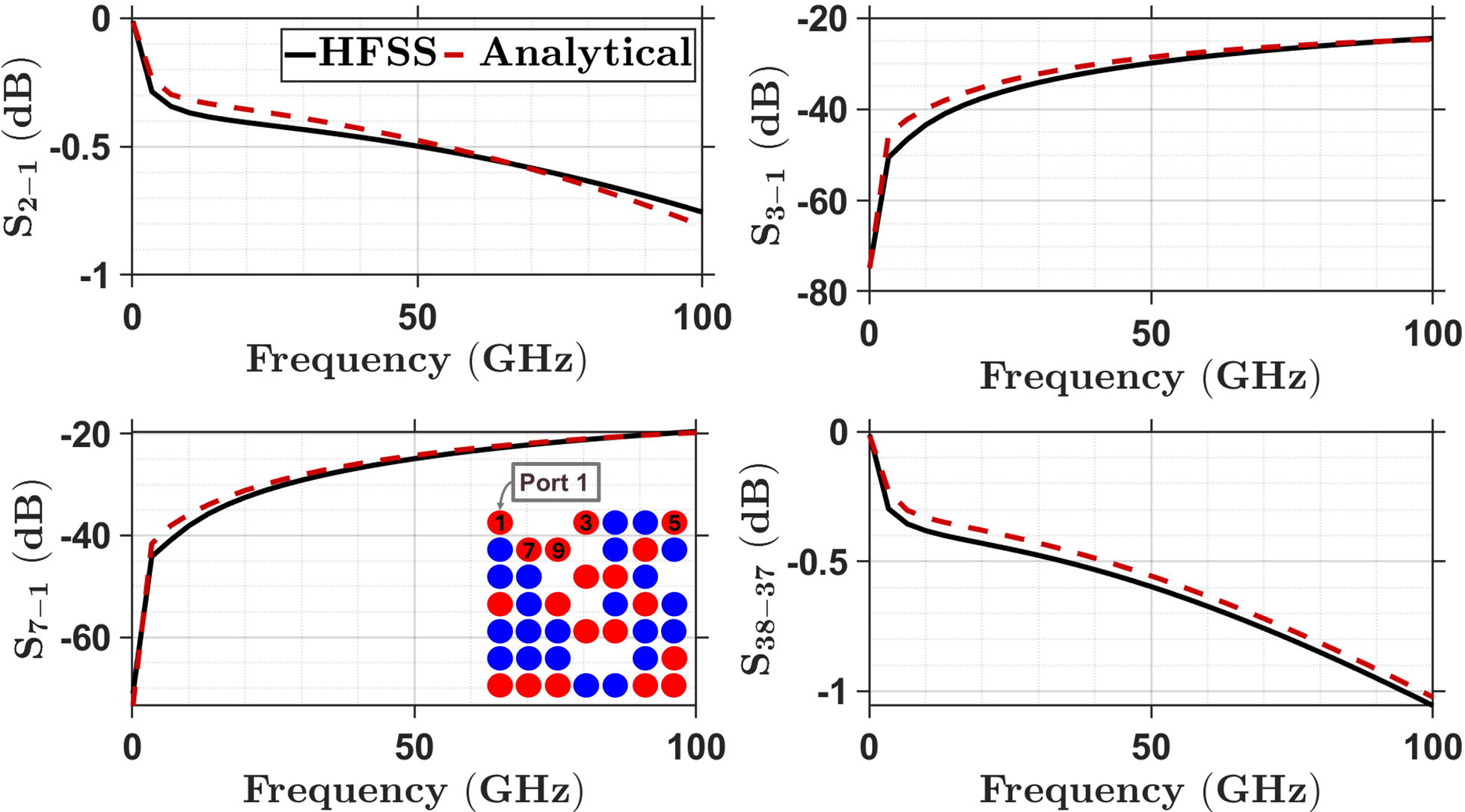}
    \caption{Magnitude comparison (in dB) of selected S-parameters obtained from HFSS full-wave simulations and the proposed analytical model. The corresponding TSV array top view is illustrated in the inset, where red circles denote signal TSVs and blue circles denote ground TSVs, arranged on a uniform pitch grid.}
    \vspace{-5pt}
    \label{fig:S_params_ana}
    \vspace{-10pt}
\end{figure}

By contrast, HFSS full-wave simulation of the same structures requires $371.8$~s for $4\times4$ and $3{,}835$~s for $25\times25$, giving the analytical solver a $\approx 13{,}700\times$ speed-up. Fig.~\ref{fig:S_params_ana} exemplifies the close agreement with HFSS on a $7\times7$ multi-ground array ($p=35$~\textmu m, $r_{\text{via}}=2.5$~\textmu m, $t_{\text{ox}}=0.25$~\textmu m, $h_{\text{via}}=100$~\textmu m, $0$--$100$~GHz). \rev{Across the $4\times4$ to $15\times15$ benchmark sweep, the analytical solver attains a complex-domain RFE (Eq.~\ref{eq:RFE}) of $5\%$--$10\%$ relative to HFSS, justifying its use as the large-scale pretraining source for Section~\ref{ML_SECTION_EXPLAIN}.}

% The same $7\times7$ array is used for thermal validation. Exciting one TSV port with $100~\text{mW}$ input power results in $3.633~\text{mW}$ of dissipated power at 98 GHz, corresponding to Joule heating computed from the S-parameters via (\ref{pinminuspout}). Two steady-state thermal simulations are performed. In the first, electromagnetic losses are exported directly from HFSS to the thermal solver and simulated using the full geometric array. In the second, the total computed power loss is applied as a uniform heat-generation source to a homogenized block of identical dimensions, characterized by effective thermal conductivities of $K_x = K_y = 140.2~\text{W/mK}$ and $K_z = 147.833~\text{W/mK}$.
% %

% The resulting steady-state temperature distributions are presented in Fig.~\ref{SS_TEMP_combined_figure}. The ETC-based model closely matches the full EM--thermal co-simulation, with a temperature deviation of less than $1\%$, while achieving approximately $10\times$ reduction in computational runtime.

To validate the homogenized ETC model, steady-state thermal simulations are conducted for $5\times5$ to $10\times10$ arrays under three boundary-condition scenarios: (i) fully populated with natural convection on all external faces and adiabatic bottom; (ii) sparse under the same natural-convection conditions; and (iii) sparse with forced convection at the top surface ($h=500~\mathrm{W/m^2K}$) and adiabatic remaining faces. Validation spans $1$--$80$~GHz across TSV radii $2$--$6$~\textmu m, pitches $20$--$40$~\textmu m, heights $50$--$100$~\textmu m, and liner thicknesses $1$--$3$~\textmu m, with $100$~mW excitation per TSV; the ETC model is compared against full Ansys Mechanical for maximum steady-state temperature and runtime.
\begin{figure}[t]
    \centering
    \includegraphics[width=0.98\linewidth]{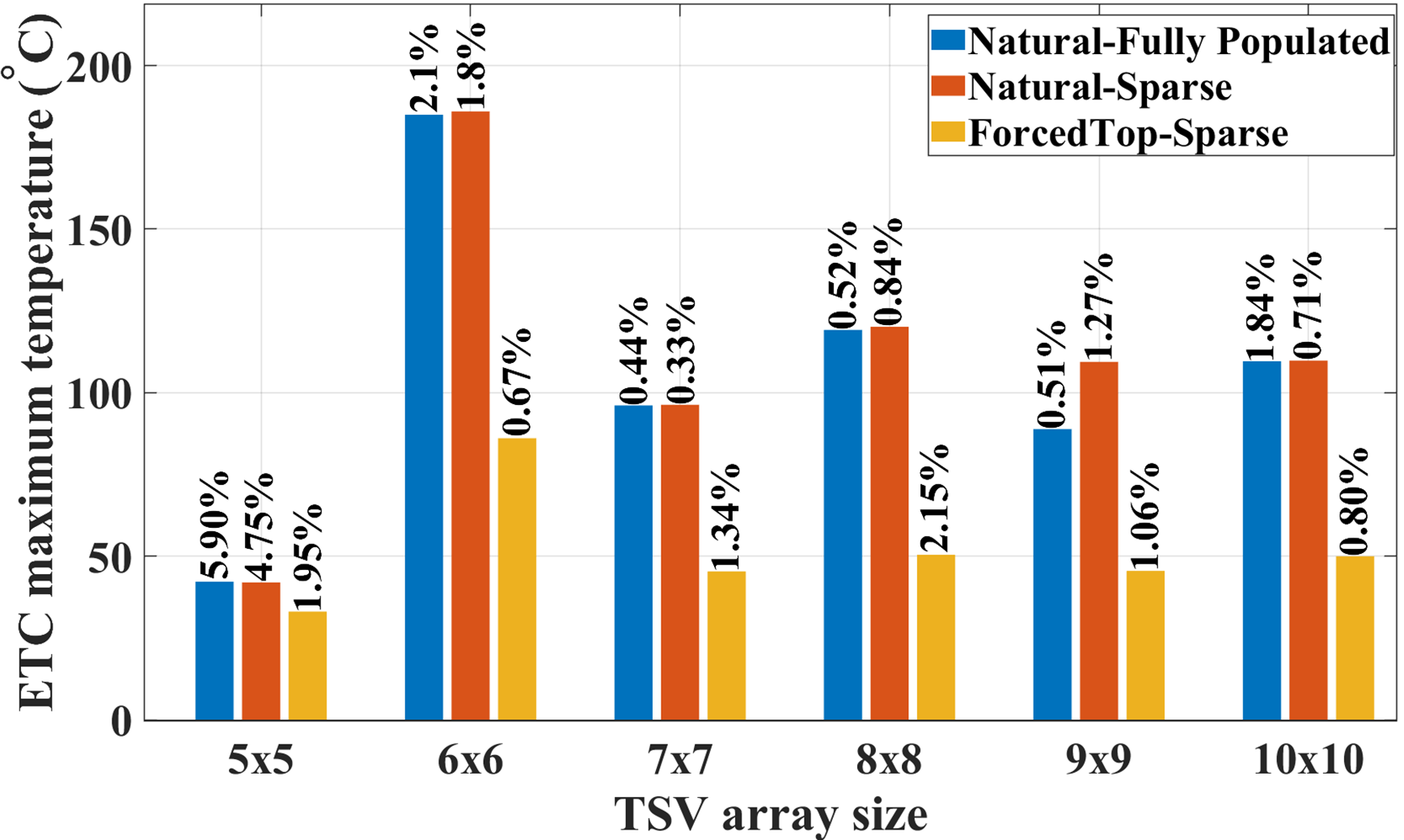}
    \caption{\rev{Comparison between the proposed homogenized ETC model and high-fidelity FEM-based Ansys Mechanical simulations for fully populated and sparse TSV arrays under natural and forced convection boundary conditions. The percentage above each bar denotes the relative error with respect to the corresponding FEM solution.}}
%    \vspace{-10pt}
    \label{fig:thermal_validation_trends}
\end{figure}

\begin{figure}[ht]
    \centering
    \vspace{-10pt}
    \includegraphics[width=0.98\linewidth]{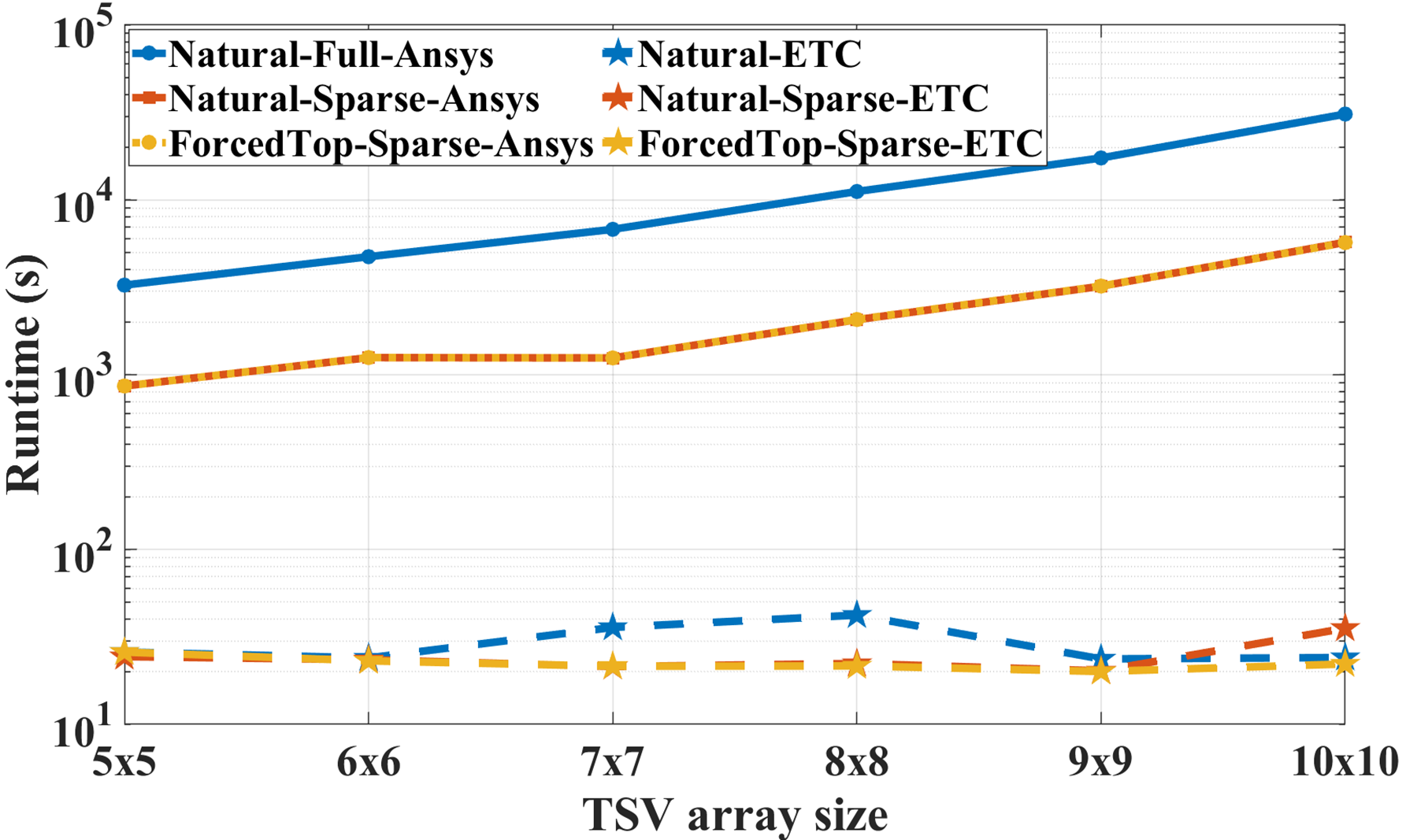}
    \caption{Total thermal analysis runtime comparison between the full detailed model and the ETC-based model.}
    \label{fig:thermal_validation_time}
\end{figure}
As shown in Fig.~\ref{fig:thermal_validation_trends}, the ETC model closely tracks the maximum temperature across array size, boundary conditions, and TSV population density: forced convection yields lower peak temperatures than the natural-convection cases, and sparse arrays run slightly hotter than fully populated ones because removing copper TSVs reduces the effective vertical conductivity. The runtime comparison in Fig.~\ref{fig:thermal_validation_time} shows speed-ups from $\approx 35\times$ for $5\times5$ sparse arrays to $\approx 1{,}300\times$ for fully populated $10\times10$, growing with array size because the full model scales rapidly with TSV count while the homogenized representation does not. Overall, the ETC model achieves a $\sim 5\%$ worst-case temperature error at orders-of-magnitude lower cost.

\subsection{ML Model Runtime Performance}

The proposed ML model is evaluated using the same dataset employed for runtime benchmarking of the analytical model, with the analysis further extended to TSV arrays ranging from $4\times4$ to $25\times25$.
As shown in Fig. \ref{fig:Inf_time}, the average inference time increases from 0.07~ms for a $4\times4$ array to 2~ms for a $15\times15$ array, and then to 18.2~ms for a $25\times25$ array, demonstrating excellent scalability with array size. Compared to full-wave simulations in HFSS, the ML model achieves a speed-up of up to $1\times10^{6}$ to $5\times10^{6}$ depending on the array size.
This substantial acceleration enables rapid exploration of extremely large design spaces, reducing evaluations that would otherwise require hours or days of FEM computation to execution times on the order of seconds.

\subsection{\texorpdfstring{\rev{Model Evaluation}}{Model Evaluation}}
\label{RFE}

\rev{The proposed surrogate is evaluated along five complementary axes: a global complex-domain agreement metric (RFE), an extrapolation analysis along four held-out axes that separates interpolation from extrapolation, a per-type signal-integrity metric suite that exposes the local worst-case errors that a single global RFE may mask, a baseline comparison against non-physics-informed surrogates, and a per-component ablation of the physics-informed architecture. The same RFE metric is used to validate the analytical solver in Section~\ref{section_analyt}.}

\subsubsection{\rev{RFE Definition and Frequency-Resolved Behavior}}
The Relative Frobenius Error (RFE) provides a compact, complex-domain measure of agreement between a predicted S-matrix and the HFSS reference, capturing both magnitude and phase discrepancies. For a predicted matrix $\mathbf{S}_{\text{model}}$ and the full-wave reference $\mathbf{S}_{\text{HFSS}}$ evaluated over all frequencies and port pairs,
\begin{equation}
\mathrm{RFE} =
\frac{\left\| \mathbf{S}_{\text{model}} - \mathbf{S}_{\text{HFSS}} \right\|_F}{\left\| \mathbf{S}_{\text{HFSS}} \right\|_F},
\label{eq:RFE}
\end{equation}
where $\|\mathbf{A}\|_F=\sqrt{\sum_{f,i,j}|A_{ij}(f)|^2}$. The normalization makes the metric scale-invariant across array sizes and frequencies. \rev{A single global RFE may average out localized errors at critical frequencies; Table~\ref{tab:r3_2_freq} resolves the per-split mean RFE across the bands $<\!1$, $1$--$10$, $10$--$30$, $30$--$60$, and $60$--$100$~GHz, confirming that the error is dominated by the highest band ($\approx 0.084$, $0.087$, $0.128$ on the three splits) while the low-to-mid bands remain at the $10^{-2}$ level even on the out-of-distribution $16{\times}16$--$25{\times}25$ split.}

\begin{table}[h]
\centering
\caption{\rev{Mean RFE by frequency band per validation split (fine-tuned reference model). The breakdown complements the global RFE by exposing frequency-localized error trends.}}
\label{tab:r3_2_freq}
\small
\setlength{\tabcolsep}{3pt} 
\begin{adjustbox}{max width=\linewidth}
\begin{tabular}{@{}lrrrrr@{}}
\toprule
split & $<\!1$\,GHz & $1$--$10$\,GHz & $10$--$30$\,GHz & $30$--$60$\,GHz & $60$--$100$\,GHz \\
\midrule
$3{\times}3$--$7{\times}7$       & 0.0110 & 0.0088 & 0.0118 & 0.0269 & 0.0838 \\
$8{\times}8$--$15{\times}15$     & 0.0168 & 0.0159 & 0.0202 & 0.0277 & 0.0868 \\
$16{\times}16$--$25{\times}25$   & 0.0566 & 0.0611 & 0.0777 & 0.0775 & 0.1281 \\
\bottomrule
\end{tabular}
\end{adjustbox}
\end{table}

\rev{
\subsubsection{Stratified Generalization}
To distinguish interpolation from extrapolation, the validation set is partitioned by array size, frequency, signal-density ratio, and geometry. As shown in Table~\ref{tab:r3_1_heldout}, size extrapolation causes a gradual RFE increase from $0.0451$ for the in-bucket $3{\times}3$--$7{\times}7$ reference to $0.0914$ for held-out $16{\times}16$--$25{\times}25$ arrays. Among the non-size axes, signal density is the most challenging, while geometry shows moderate degradation and frequency remains comparable to the in-region reference except for the largest arrays. These trends indicate that dense signaling and large-array regimes dominate the extrapolation difficulty, although the surrogate remains suitable for design-space ranking.

To assess the analytical prior, Table~\ref{tab:r3_1_pft} compares analytical pre-training only, HFSS-only training, and analytical pre-training followed by HFSS fine-tuning. HFSS-only training gives the best in-distribution accuracy, whereas the two-stage pre-train/fine-tune regime gives the best out-of-distribution (OOD) performance. For the $16{\times}16$--$25{\times}25$ split, the mean RFE improves from $0.1824$ to $0.0982$ and $0.0914$, respectively, confirming that the OOD gain comes from transfer learning rather than the analytical prior alone. Across held-out slices, the RFE remains in the $10^{-2}$--$10^{-1}$ range, with Kendall's $\tau\ge 0.78$ across all splits (Table~\ref{tab:r3_2_suite}).
}

\begin{table}[t]
\centering
\caption{\rev{Three-regime training ablation on identical HFSS splits. Regimes: (a) analytical pre-train only; (b) HFSS-only from scratch; (c) analytical pre-train $+$ HFSS fine-tune. Splits given as $N{\times}N$ bucket range. Per-split means.}}
\label{tab:r3_1_pft}
\small
\setlength{\tabcolsep}{4pt}
\begin{adjustbox}{max width=\linewidth}
\begin{tabular}{@{}clrrrr@{}}
\toprule
split & regime & RFE & \makecell[r]{MAE\\NEXT} & \makecell[r]{MAE\\FEXT} & \makecell[r]{Worst victim\\NEXT (dB)} \\
\midrule
\multirow{3}{*}{3--7}
  & (a) & 0.0813 & 0.0096 & 0.0086 & 1.75 \\
  & (b) & 0.0288 & 0.0045 & 0.0041 & 0.89 \\
  & (c) & 0.0451 & 0.0057 & 0.0055 & 1.05 \\
\addlinespace
\multirow{3}{*}{8--15}
  & (a) & 0.0878 & 0.0068 & 0.0056 & 2.72 \\
  & (b) & 0.0331 & 0.0023 & 0.0025 & 1.07 \\
  & (c) & 0.0488 & 0.0034 & 0.0031 & 1.33 \\
\addlinespace
\multirow{3}{*}{16--25}
  & (a) & 0.1824 & 0.0097 & 0.0070 & 4.30 \\
  & (b) & 0.0982 & 0.0030 & 0.0055 & 3.19 \\
  & (c) & 0.0914 & 0.0039 & 0.0041 & 3.81 \\
\bottomrule
\end{tabular}
\end{adjustbox}
\end{table}

\begin{table}[t]
\centering
\caption{\rev{Held-out-slice generalization: mean RFE on each held-out slice versus matched in-region reference. Size buckets given as $N{\times}N$ range. Size rows reuse Table~\ref{tab:r3_1_pft} regime (c); other rows train on the complement.}}
\label{tab:r3_1_heldout}
\small
\setlength{\tabcolsep}{4pt}
\begin{adjustbox}{max width=\linewidth}
\begin{tabular}{@{}llrr@{}}
\toprule
slice & size & RFE held & \makecell[r]{RFE in-region\\(3--7 ref.)} \\
\midrule
\multirow{3}{*}{array size}
  & 3--7   & 0.0451 & 0.0451 \\
  & 8--15  & 0.0488 & 0.0451 \\
  & 16--25 & 0.0914 & 0.0451 \\
\addlinespace
\multirow{3}{*}{freq.\ $30$--$60$\,GHz}
  & 3--7   & 0.0286 & 0.0416 \\
  & 8--15  & 0.0518 & 0.0587 \\
  & 16--25 & 0.1627 & 0.1114 \\
\addlinespace
\multirow{3}{*}{signal density $>\!0.55$}
  & 3--7   & 0.0628 & 0.0248 \\
  & 8--15  & 0.0787 & 0.0517 \\
  & 16--25 & 0.1651 & 0.1082 \\
\addlinespace
\multirow{3}{*}{\makecell[l]{geom.\ $R>5\,\mu$m,\\$P<30\,\mu$m}}
  & 3--7   & 0.0395 & 0.0368 \\
  & 8--15  & 0.0529 & 0.0526 \\
  & 16--25 & 0.1339 & 0.1075 \\
\bottomrule
\end{tabular}
\end{adjustbox}
\end{table}

\rev{
\subsubsection{Signal-Integrity Metric Suite}
To complement global RFE, we evaluate a signal-integrity metric suite that captures local worst-case behavior, including per-type MAE for $S_{21}$, $S_{11}$, NEXT, and FEXT; worst-victim NEXT/FEXT dB error at the largest-coupling pair; dominant-pair dB error at the strongest true $|S|$ entry; and Kendall's $\tau$ rank correlation for NEXT/FEXT-based design ranking. dB errors are evaluated only on dominant coupling terms and floored at $-60$\,dB, consistent with the practical HFSS de-embedding noise floor.

As shown in Table~\ref{tab:r3_2_suite}, the mean RFE remains below $5\%$ in-distribution and below $10\%$ on the OOD $16{\times}16$--$25{\times}25$ split. Per-type MAEs stay at the $10^{-3}$ level, worst-victim NEXT/FEXT errors remain within $\sim$1--2\,dB in-distribution, the dominant-pair dB error stays below $0.4$\,dB, and Kendall's $\tau$ remains at least $0.85$ in-distribution and $0.78$ OOD.
}

\begin{table*}[t]
\centering
\caption{\rev{Combined per-split signal-integrity metric suite (left block) and ranking-fidelity (right block) for the fine-tuned reference model. ``Worst victim NEXT/FEXT (dB)'' = absolute dB error at the largest-coupling NEXT (resp.\ FEXT) pair; ``dom-pair dB'' = absolute dB error at the matrix pair with the largest true $|S|$, which is typically the dominant transmission term; $\tau$~=~Kendall's rank correlation (HFSS vs surrogate). All values are means; dB metrics floored at $-60$\,dB.}}
\label{tab:r3_2_suite}
\small
\setlength{\tabcolsep}{4pt}
\begin{adjustbox}{max width=\textwidth}
\begin{tabular}{@{}lrrrrrrrr|rrr@{}}
\toprule
split & \makecell[r]{RFE\\mean} & \makecell[r]{MAE\\$S_{21}$} & \makecell[r]{MAE\\$S_{11}$} & \makecell[r]{MAE\\NEXT} & \makecell[r]{MAE\\FEXT} & \makecell[r]{Worst victim\\NEXT (dB)} & \makecell[r]{Worst victim\\FEXT (dB)} & \makecell[r]{dom-pair\\dB} & $\tau_{\mathrm{NEXT}}$ & $\tau_{\mathrm{FEXT}}$ & $\tau_{\mathrm{FOM}}$ \\
\midrule
$3{\times}3$--$7{\times}7$       & 0.0451 & 0.0207 & 0.0097 & 0.0057 & 0.0055 & 1.05 & 1.43 & 0.23 & 0.908 & 0.899 & 0.909 \\
$8{\times}8$--$15{\times}15$     & 0.0488 & 0.0112 & 0.0085 & 0.0034 & 0.0031 & 1.33 & 2.06 & 0.10 & 0.870 & 0.852 & 0.871 \\
$16{\times}16$--$25{\times}25$   & 0.0914 & 0.0143 & 0.0093 & 0.0039 & 0.0041 & 3.81 & 5.37 & 0.38 & 0.810 & 0.784 & 0.803 \\
\bottomrule
\end{tabular}
\end{adjustbox}
\end{table*}

\rev{
\subsubsection{Baseline Comparison}
To isolate the effect of the physics-informed components, the proposed surrogate is compared with three non-physics-informed baselines under the same HFSS dataset, splits, seed, and early-stopping protocol: PlainGNN, an MLP using aggregated design features, and XGBoost using the same aggregate features. PlainGNN uses the same graph backbone but removes FiLM conditioning, dual heads, reciprocity averaging, and the passivity penalty, while the MLP and XGBoost predict only aggregate $8$-D targets rather than the full pair-wise S-matrix.

As shown in Table~\ref{tab:r1_4_baselines}, both the proposed model and PlainGNN achieve $10^{-3}$-level per-type MAE and sub-$0.1$ RFE on the in-distribution splits. However, on the OOD $16{\times}16$--$25{\times}25$ split, the proposed model degrades less than PlainGNN, reducing RFE from $0.1486$ to $0.0982$ and maintaining stronger ranking fidelity. In contrast, the MLP collapses on larger arrays, while XGBoost remains limited to aggregate predictions and cannot provide per-pair S-matrix or worst-victim metrics. The proposed model also maintains $\tau_{\mathrm{FOM}}\ge 0.82$ across all splits, compared with $0.636$ for PlainGNN and $0.121$ for the MLP on the OOD bucket. On the largest split, its per-output inference cost is only modestly higher than PlainGNN, but remains substantially lower than the MLP and XGBoost, demonstrating that the physics-informed terms improve extrapolation and ranking with limited computational overhead.
}

\begin{table}[t]
\centering
\caption{\rev{Matched-protocol comparison with non-physics-informed baselines on the same $9\,000$-design HFSS set. PlainGNN removes the proposed physics-informed terms, while MLP/XGBoost use aggregate features and targets. $\tau_{\mathrm{FOM}}$ is Kendall's rank correlation on $\max(|\mathrm{NEXT}|,|\mathrm{FEXT}|)$; cost is reported per output on the largest split. The ``Proposed'' row is trained from scratch and therefore differs from the pre-train/fine-tune reference in Tables~\ref{tab:r3_1_pft}--\ref{tab:r3_2_suite}.}}

\label{tab:r1_4_baselines}
\small
\setlength{\tabcolsep}{3pt}
\begin{adjustbox}{max width=\linewidth}
\begin{tabular}{@{}llrrrrrr@{}}
\toprule
model & split & \makecell[r]{MAE\\NEXT} & \makecell[r]{MAE\\FEXT} & RFE & \makecell[r]{Worst victim\\NEXT (dB)} & $\tau_{\text{FOM}}$ & \makecell[r]{ns / output\\($16$--$25$)} \\
\midrule
\multirow{3}{*}{Proposed}
  & 3--7   & 0.0045 & 0.0041 & 0.0288 & 0.89 & 0.933 & \multirow{3}{*}{26.5} \\
  & 8--15  & 0.0023 & 0.0025 & 0.0331 & 1.07 & 0.910 & \\
  & 16--25 & 0.0030 & 0.0055 & 0.0982 & 3.19 & 0.821 & \\
\addlinespace
\multirow{3}{*}{PlainGNN}
  & 3--7   & 0.0043 & 0.0040 & 0.0246 & 1.04 & 0.925 & \multirow{3}{*}{22.2} \\
  & 8--15  & 0.0036 & 0.0040 & 0.0513 & 1.27 & 0.882 & \\
  & 16--25 & 0.0052 & 0.0073 & 0.1486 & 3.97 & 0.636 & \\
\addlinespace
\multirow{3}{*}{MLP}
  & 3--7   & 0.0257 & 0.0242 & --- & --- & 0.670 & \multirow{3}{*}{9\,634} \\
  & 8--15  & 0.0978 & 0.3019 & --- & --- & 0.298 & \\
  & 16--25 & 3.3027 & 2.3408 & --- & --- & 0.121 & \\
\addlinespace
\multirow{3}{*}{XGBoost}
  & 3--7   & 0.0241 & 0.0220 & --- & --- & 0.693 & \multirow{3}{*}{41\,551} \\
  & 8--15  & 0.0175 & 0.0170 & --- & --- & 0.782 & \\
  & 16--25 & 0.0144 & 0.0146 & --- & --- & 0.845 & \\
\bottomrule
\end{tabular}
\end{adjustbox}
\end{table}

\rev{
\subsubsection{Ablation Studies}
Each physics-informed component is evaluated under a matched protocol using the same $9{,}000$-design HFSS set, hyperparameters, seed, and early-stopping split. The reference model includes FiLM conditioning, reciprocity symmetrization, and a passivity penalty $\lambda=0.05$; the ablations remove one component at a time.

As shown in Table~\ref{tab:r3_6_ablation}, FiLM consistently outperforms parameter-matched concatenation conditioning across all splits, reducing RFE from $0.0626/0.1129/0.1505$ to $0.0288/0.0331/0.0982$. Reciprocity symmetrization enforces zero reciprocity residual by construction and further reduces RFE on all splits. The passivity penalty improves in-distribution accuracy relative to the unconstrained model and remains essentially tied on the OOD split, indicating that enforcing $\sigma_{\max}(\mathbf{S})\le 1$ improves physical consistency without degrading extrapolation.
}

\begin{table}[t]
\centering
\caption{\rev{Architecture ablation (matched protocol; HFSS-only, identical hyperparameters and seed). ``Reference'' = full model (FiLM, reciprocity averaging, passivity penalty $\lambda=0.05$); each other arm removes one component. The validation-split label ``$N_1$--$N_2$'' denotes the array-size bucket from $N_1{\times}N_1$ through $N_2{\times}N_2$. ``Worst victim NEXT (dB)'' = absolute dB error at the largest-coupling NEXT pair; ``recip residual'' = mean $|S_{ij}-S_{ji}|$ on the raw edge head.}}
\label{tab:r3_6_ablation}
\footnotesize
\setlength{\tabcolsep}{3pt}
\renewcommand{\arraystretch}{1.1}
\begin{tabular}{@{}p{1.75cm}crrrr@{}}
\toprule
arm & split & RFE & \makecell[r]{MAE\\NEXT} & \makecell[r]{Worst victim\\NEXT (dB)} & \makecell[r]{recip\\residual} \\
\midrule
\multirow{3}{1.75cm}{\raggedright Reference (full proposed model)}
  & 3--7   & 0.0288 & 0.0045 & 0.89 & 0.000000 \\
  & 8--15  & 0.0331 & 0.0023 & 1.07 & 0.000000 \\
  & 16--25 & 0.0982 & 0.0030 & 3.19 & 0.000000 \\
\addlinespace
\multirow{3}{1.75cm}{\raggedright w/o FiLM (concatenation)}
  & 3--7   & 0.0626 & 0.0092 & 1.53 & 0.000000 \\
  & 8--15  & 0.1129 & 0.0087 & 3.01 & 0.000000 \\
  & 16--25 & 0.1505 & 0.0060 & 6.52 & 0.000000 \\
\addlinespace
\multirow{3}{1.75cm}{\raggedright w/o reciprocity (forward only)}
  & 3--7   & 0.0509 & 0.0076 & 1.61 & 0.000203 \\
  & 8--15  & 0.0821 & 0.0050 & 2.36 & 0.000203 \\
  & 16--25 & 0.1308 & 0.0051 & 5.17 & 0.000203 \\
\addlinespace
\multirow{3}{1.75cm}{\raggedright w/o passivity penalty ($\lambda=0$)}
  & 3--7   & 0.0335 & 0.0052 & 1.18 & 0.000000 \\
  & 8--15  & 0.0474 & 0.0034 & 1.85 & 0.000000 \\
  & 16--25 & 0.0969 & 0.0037 & 4.07 & 0.000000 \\
\bottomrule
\end{tabular}
\end{table}

\rev{
\subsection{Coupled Electrothermal Evaluation}
Using the unified framework in Section~\ref{Unified_section}, a $4{\times}4$ TSV array is evaluated in a $300{\times}300{\times}80~\mu\mathrm{m}$ silicon substrate with a $50{\times}50{\times}10~\mu\mathrm{m}$ temperature-dependent silicon die heat source bonded at one corner. The array contains 15 signal TSVs and one ground TSV, with $r_{\mathrm{via}}=2~\mu\mathrm{m}$, $h_{\mathrm{via}}=80~\mu\mathrm{m}$, $p=25~\mu\mathrm{m}$, and $t_{\mathrm{ox}}=1.5~\mu\mathrm{m}$. The die power follows $P_{\mathrm{die}}(T_{\mathrm{die}})=P_{0}+\alpha_{\mathrm{die}}(T_{\mathrm{die}}-T_{\mathrm{ref}})$, with $P_{0}=0.3~\mathrm{W}$ and $\alpha_{\mathrm{die}}=5~\mathrm{mW/K}$, to model first-order CMOS leakage feedback~\cite{7927273}.}

\rev{The thermal boundary conditions use $T_{\infty}=300~\mathrm{K}$, with a bottom cold plate ($h=4\times10^{5}~\mathrm{W/m^{2}\!\cdot\!K}$), natural convection on the sidewalls, and weak convection on the top surfaces. Electrically, all signal TSV top ports are driven at $f_{0}=100~\mathrm{GHz}$ with $P_{\mathrm{in}}=1~\mathrm{W}$ per port, giving $15~\mathrm{W}$ total incident power. The fixed-point electrothermal iteration is stopped when}
\begin{equation}
\left|
\overline{T}_{\mathrm{sub}}^{(k+1)}
-
\overline{T}_{\mathrm{sub}}^{(k)}
\right|
\le
0.5~\mathrm{K}.
\end{equation}

\rev{The proposed GNN+ETC flow converges in $14$ iterations to
$\overline{T}_{\mathrm{sub}}=350.0\,\mathrm{K}$,
$T_{\mathrm{die}}=545.5\,\mathrm{K}$, and
$P_{\mathrm{die}}=1.515\,\mathrm{W}$. Compared with the fully
bi-directional Ansys HFSS$\leftrightarrow$Mechanical reference, which
converges in $15$ iterations to $367.0\,\mathrm{K}$, $578.5\,\mathrm{K}$,
and $1.682\,\mathrm{W}$, the GNN+ETC errors are
$4.6\%$, $5.7\%$, and $9.9\%$, respectively. The Ansys reference maps
non-uniform per-TSV HFSS EM losses into Mechanical and feeds the resolved
per-TSV temperatures back to update $\sigma_{\mathrm{Cu}}(T)$ in the next
HFSS solve.}

\rev{To identify the source of the residual error and validate the full coupled analytical model, the GNN electrical step
is replaced by an analytical HSPICE model coupled to the same ETC thermal
solver. This HSPICE+ETC flow converges in $15$ iterations to
$\overline{T}_{\mathrm{sub}}=348.3\,\mathrm{K}$,
$T_{\mathrm{die}}=541.0\,\mathrm{K}$, and
$P_{\mathrm{die}}=1.494\,\mathrm{W}$, corresponding to
$5.1\%/6.5\%/11.2\%$ undershoot relative to Ansys. Since both surrogate
pipelines undershoot Ansys with the same sign and similar magnitude, the
remaining discrepancy is mainly attributed to the shared homogenized ETC
thermal approximation rather than the electrical surrogate. The
three-way iteration histories and convergence trends are shown in
Fig.~\ref{fig:r13_aligned_et_overlay_ms}.}

\rev{
\begin{figure}[t]
    \centering
    \includegraphics[width=0.98\linewidth]{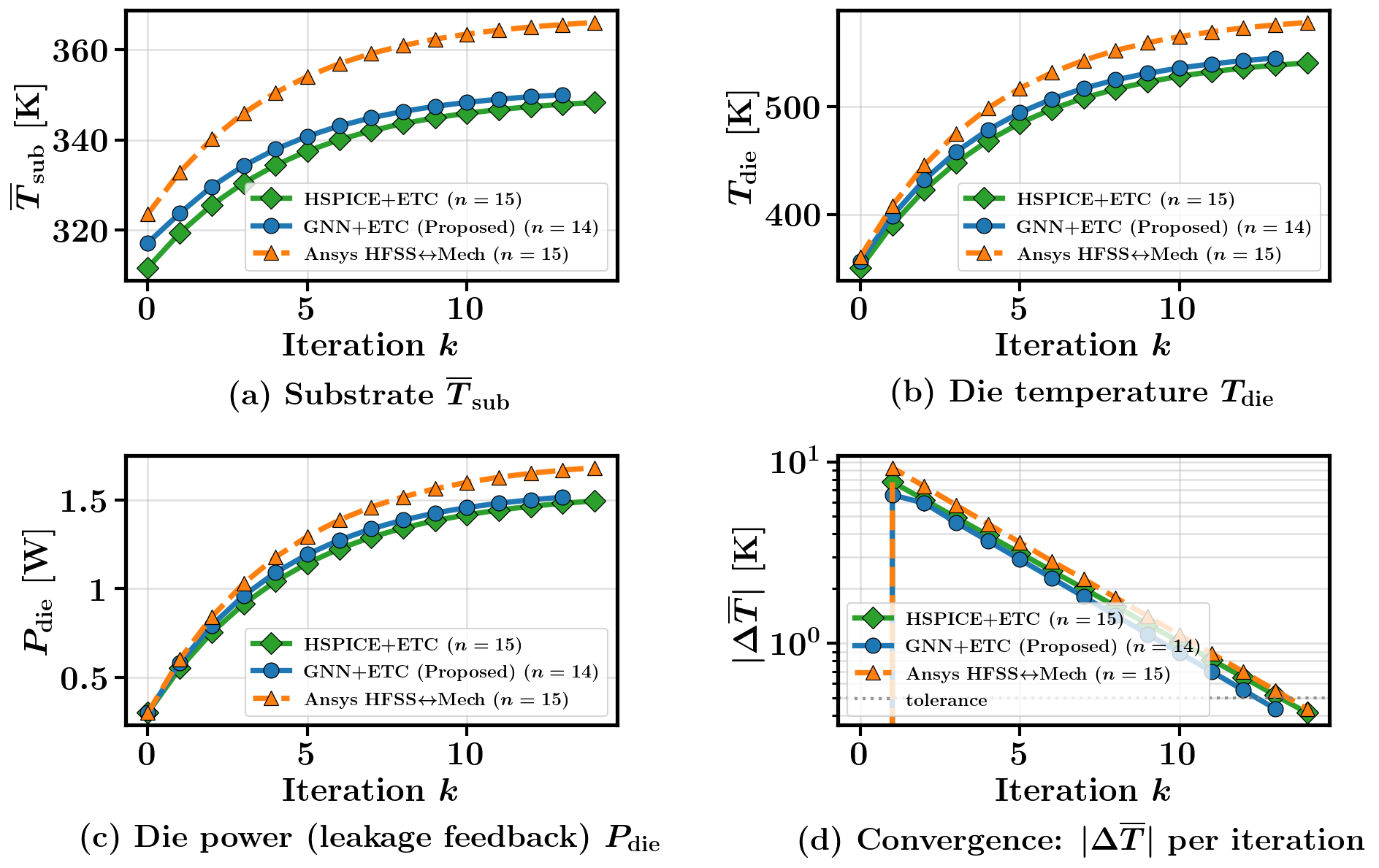}
\caption{\rev{Coupled electrothermal trajectories for the representative $4{\times}4$ testcase. GNN+ETC, HSPICE+ETC, and the Ansys HFSS$\leftrightarrow$Mechanical reference are compared for: (a) $\overline{T}_{\mathrm{sub}}$, (b) $T_{\mathrm{die}}$, (c) $P_{\mathrm{die}}(T)$, and (d) $|\Delta\overline{T}|$ convergence.}}

    \label{fig:r13_aligned_et_overlay_ms}
\end{figure}
}
\vspace{-10pt}
\subsection{ML Framework for Optimization}

\begin{table}[t]
\centering
\caption{TSV Array Physical and Geometrical Parameters.}
\label{tab:tsv_parameters}
\begin{tabular}{ccc}
\hline\hline
\textbf{Parameters} & \textbf{Symbol} & \textbf{Value} \\
\hline
TSV radius & $r_{\mathrm{TSV}}$ & $5~\mu\mathrm{m}$ \\
TSV height & $h_{\mathrm{TSV}}$ & $100~\mu\mathrm{m}$ \\
Pitch between adjacent TSVs & $p$ & $60~\mu\mathrm{m}$ \\
Thickness of SiO$_2$ layer & $t_{\mathrm{ox}}$ & $0.5~\mu\mathrm{m}$ \\
Copper conductivity & $\sigma_{\mathrm{Cu}}$ & $5.8\times10^{7}~\mathrm{S/m}$ \\
Silicon conductivity & $\sigma_{\mathrm{Si}}$ & $10~\mathrm{S/m}$ \\
Dielectric constant of SiO$_2$ & $\varepsilon_{\mathrm{ox}}$ & $4$ \\
\hline\hline
\end{tabular}
\end{table}

In addition to serving as a fast surrogate model for S-parameter prediction, the proposed ML framework can be directly leveraged as an efficient optimization engine for TSV array design. Specifically, once trained, the ML model enables rapid evaluation of a large combinatorial design space without requiring repeated full-wave simulations.

% To demonstrate the optimization capability of the proposed framework, the benchmark problem introduced in~\cite{10830538} is revisited. In that work, a $5\times5$ TSV array comprising 12 signal TSVs and 13 ground TSVs is considered, with all physical and geometrical parameters fixed as summarized in Table~\ref{tab:tsv_parameters}. The optimization objective is to determine the optimal placement of the signal TSVs that minimizes the overall crosstalk within the array.

\rev{To demonstrate the optimization capability of the proposed framework, the benchmark problem introduced in~\cite{10830538} is revisited. The benchmark consists of a $5\times5$ TSV array with 12 signal TSVs and 13 ground TSVs. To ensure a consistent comparison, the physical dimensions, material parameters, operating frequency of 15~GHz, single-ended port assignment with top and bottom ports for each TSV, $50~\Omega$ terminations, and signal count are kept identical to the baseline setup. The corresponding physical and geometrical parameters are summarized in Table~\ref{tab:tsv_parameters}. The optimization task is to determine the signal/ground placement that reduces crosstalk among the signal TSVs.}
\begin{figure*}[t]
    \centering
    
    \begin{subfigure}[t]{0.49\textwidth}
        \centering
        \includegraphics[width=\linewidth]{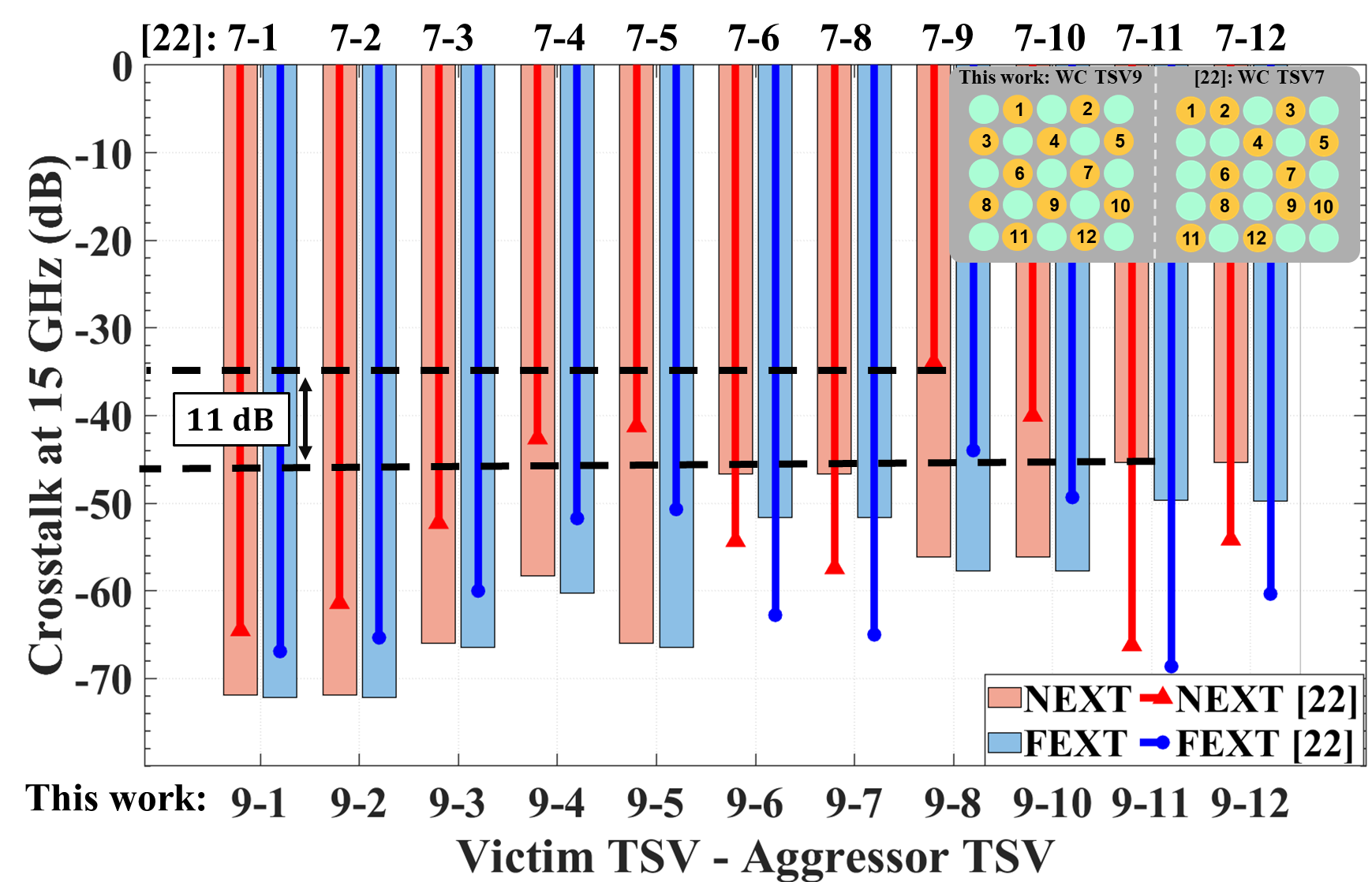}
        \caption{}
        \label{fig:xtalk_12_sig}
    \end{subfigure}
    \hfill
    \begin{subfigure}[t]{0.49\textwidth}
        \centering
        \includegraphics[width=\linewidth]{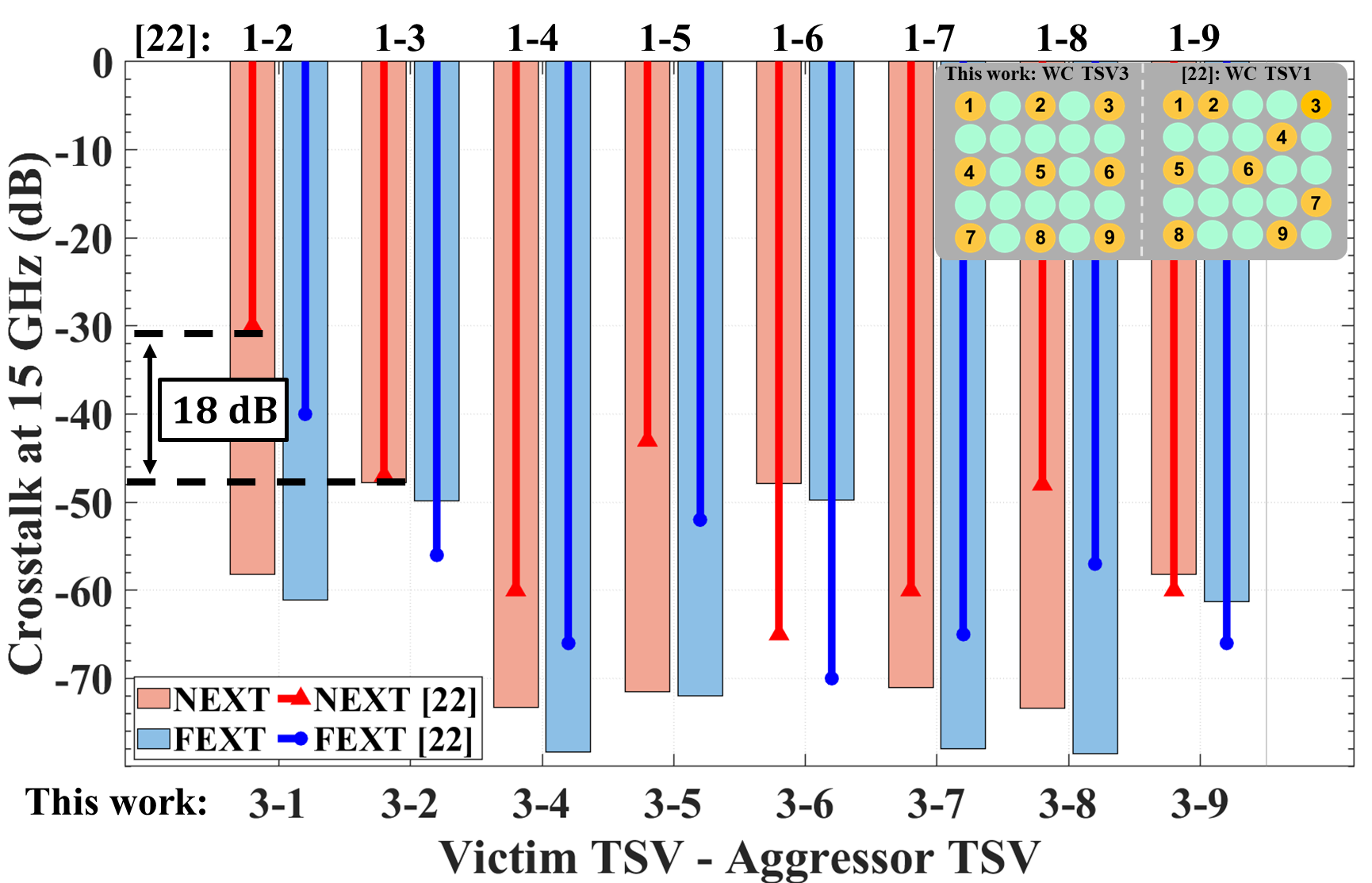}
        \caption{}
        \label{fig:xtalk_9_sig}
    \end{subfigure}
    
    \caption{Near-end (NEXT) and far-end (FEXT) crosstalk at 15~GHz for the worst-case victim TSV, comparing the proposed signal--ground assignment with the state-of-the-art configuration in~\cite{10830538}. Insets show the corresponding TSV layouts. Bars denote the proposed design and line markers the results reported in~\cite{10830538}. The horizontal dashed lines indicate the worst-case total crosstalk, showing an improvement of (a) 11~dB in the 12-signal ($5\times5$) array and (b) 18~dB in the 9-signal ($5\times5$) array.}
    
    \label{fig:xtalk_cases}
    \vspace{-10pt}
\end{figure*}
The performance metric used for optimization is the average crosstalk across all signal TSVs. This metric accounts for both near-end crosstalk (NEXT) and far-end crosstalk (FEXT) contributions and is defined as
\begin{equation}
X_{\mathrm{talk}} =
\frac{1}{s(s-1)}
\sum_{v=1}^{s}
\left| \mathrm{dB}\!\left( X_{v}^{\mathrm{tot}} \right) \right|,
\label{eq:xtalk}
\end{equation}
%
% \begin{equation}
% \mathrm{Xtalk} = 
% \frac{1}{s(s-1)}
% \sum_{i=1}^{s}
% \left| \mathrm{dB}\!\left(\mathrm{Total\_Xtalk}_i\right) \right|,
% \label{eq:xtalk}
% \end{equation}
%
%
where $s$ denotes the number of signal TSVs, and $X_{v}^{\mathrm{tot}}$ represents the total coupling experienced by the $v$-th victim signal TSV. \rev{Since the baseline does not explicitly specify how the individual aggressor contributions are aggregated to form $\left| \mathrm{dB}\!\left( X_{v}^{\mathrm{tot}} \right) \right|$, the aggregation used in this work is defined explicitly. For each victim TSV, the total crosstalk is computed by aggregating the magnitudes of NEXT and FEXT contributions from all other aggressor signal TSVs using an RMS-based aggregation to capture the combined energy of coupling terms:}

% \begin{equation}
% \mathrm{Total\_Xtalk}_i
% =
% \sqrt{\sum_{a\neq v}\left(M_{\text{NEXT}}^{(v,a)}\right)^2
% +
% \sum_{a\neq v}\left(M_{\text{FEXT}}^{(v,a)}\right)^2 }.
% \label{eq:total_xtalk_rms}
% \end{equation}
\begin{equation}
X_{v}^{\mathrm{tot}} =
\sqrt{
\sum_{\substack{a=1 \\ a \neq v}}^{s}
\left( M_{\mathrm{NEXT}}^{(v,a)} \right)^{2}
+
\sum_{\substack{a=1 \\ a \neq v}}^{s}
\left( M_{\mathrm{FEXT}}^{(v,a)} \right)^{2}
},
\label{eq:total_xtalk_rms}
\end{equation}
where $v$ denotes the index of the victim signal TSV, and $a$ denotes the index of an aggressor signal TSV. The terms $M_{\text{NEXT}}^{(v,a)}$ and $M_{\text{FEXT}}^{(v,a)}$ represent the magnitudes of, respectively, the near-end and far-end crosstalk components from aggressor $a$ to victim $v$. This formulation ensures that both magnitude and distribution of coupling contributions are properly reflected in the victim's total interference level \rev{and is used only as the internal optimization objective.}
\rev{For a strict comparison with~\cite{10830538}, the final reported improvement is not taken from the aggregate objective in~(\ref{eq:xtalk}). After the optimized TSV placement is obtained, the selected layout is evaluated using the same worst-case crosstalk criterion reported in the baseline. Specifically, the worst victim TSV and the corresponding worst NEXT and FEXT coupling values are extracted over all victim--aggressor signal TSV pairs under the matched geometry, frequency, port assignment, termination, and signal-count conditions. The reported dB improvement is therefore based on this post-optimization worst-case NEXT/FEXT evaluation, rather than on the RMS aggregate metric used during search.}

% \begin{figure}[t]
%     \centering
%     \includegraphics[width=0.95\linewidth]{images/Xtalk_WorstVictim_TSV5_15GHz.png}
%     \caption{Near-end crosstalk (NEXT) and far-end crosstalk (FEXT) at 15~GHz for the worst-case victim TSV (TSV~5) in a $5\times5$ TSV array. The bar pairs correspond to NEXT (red) and FEXT (blue) coupling from each aggressor signal TSV (1--12) to the selected victim. The inset illustrates the physical arrangement of signal (red) and ground (blue) TSVs within the array.}
%     \vspace{-10pt}
%     \label{fig:TSV_CASE_1}
% \end{figure}
% \begin{figure}[t]
%     \centering
%     \includegraphics[width=0.95\linewidth]{images/Fig_11_VS_2.png}
%     \caption{Near-end (NEXT) and far-end (FEXT) crosstalk at 15~GHz for the worst-case victim TSV in the $5\times5$ TSV array (12 signal TSVs), comparing the proposed signal--ground assignment with the state-of-the-art configuration in~[18]. Light bars denote the proposed design and dark bars denote~[18]. The dashed lines highlight the worst-case total crosstalk levels, revealing an overall reduction of approximately 11~dB. The inset shows the signal and ground TSV distributions for both layouts (green/yellow for~[18] and blue/red for the proposed configuration).}
%     \vspace{-10pt}
%     \label{fig:TSV_CASE_1}
% \end{figure}

Under this configuration, the optimization reduces to placing 12 signal TSVs among the 25 grid locations. The total number of possible configurations is
\begin{equation}
\binom{25}{12} = 5{,}200{,}300.
\label{eq:comb_inferences}
\end{equation}
\greenrev{Exhaustive exploration with full-wave FEM is computationally prohibitive, whereas the proposed ML surrogate enables millisecond-scale inference and evaluates the entire space in approximately $5$~minutes. Further, the square TSV grid exhibits dihedral symmetry ($D_4$): if $f(x)$ denotes the surrogate mapping from a placement $x$ to its S-parameters and $g(x)$ a symmetry transformation, then $f(g(x)) = g(f(x))$ for all $x$, so configurations in the same orbit are electrically equivalent up to permutation. Evaluating one representative per orbit reduces the effective space by up to $8\times$, allowing the symmetry-reduced set ($\approx 6.5\times10^{5}$ configurations) to be exhaustively scored in under one minute and a complete Pareto frontier constructed.}

\greenrev{As shown in Fig.~\ref{fig:xtalk_12_sig}, the optimal $12$-signal arrangement yields a worst-victim crosstalk no higher than $\approx -45$~dB across all aggressor couplings, an $\sim 11$~dB improvement ($\sim 3.5\times$ linear reduction) over the $\approx -34$~dB baseline of~\cite{10830538}.}

\begin{figure}[ht]
    \centering
    \includegraphics[width=0.95\linewidth]{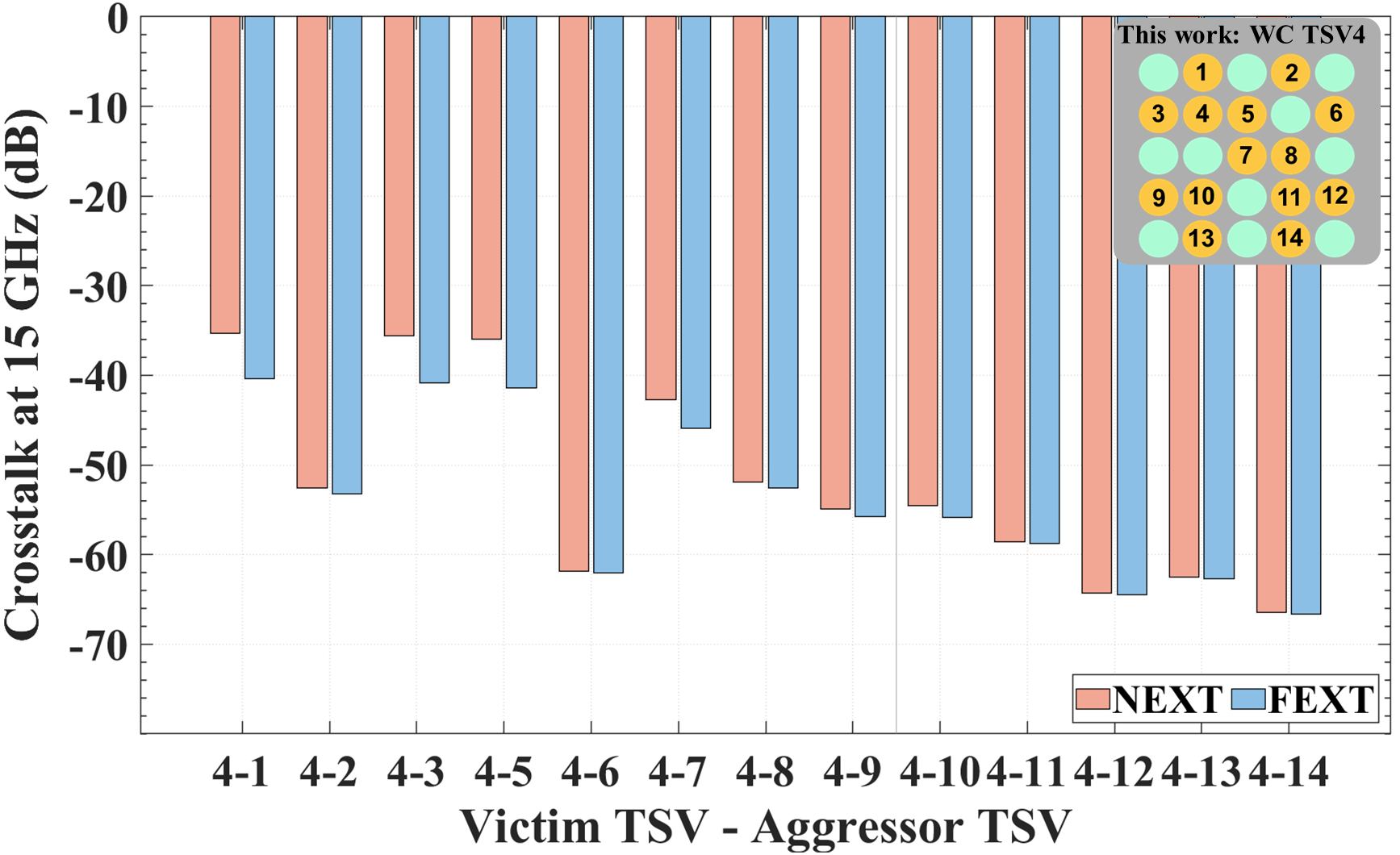}
    \caption{Results of a design space exploration 9-15 signal TSVs in a $5\times5$ array. Near-end (NEXT) and far-end (FEXT) crosstalk at 15~GHz for the worst-case victim TSV in the optimized signal--ground assignment (14 signal TSVs). The inset shows the corresponding TSV layout.}
    \label{fig:TSV_CASE_3}
\end{figure}

\greenrev{A relaxed variant of the same benchmark in~\cite{10830538} allows the signal count to vary between $9$ and $15$, with all other geometric and material parameters from Table~\ref{tab:tsv_parameters} maintained. The cumulative design space across these counts reaches $\approx 2.8\times10^{7}$ configurations (per~(\ref{eq:comb_inferences})); $D_4$ deduplication and the millisecond surrogate again make Pareto-frontier construction tractable. The optimal $9$-signal layout (Fig.~\ref{fig:xtalk_9_sig}) reaches a worst-victim crosstalk of $-48$~dB (between TSV~3 and TSV~2), an $\sim 18$~dB improvement ($\sim 8\times$ linear reduction) over~\cite{10830538}. A higher-density $14$-signal layout (Fig.~\ref{fig:TSV_CASE_3}) reaches $-35.2$~dB on the worst victim (TSV~4 vs.\ TSV~1), still $>5$~dB better than the $9$-TSV layout in~\cite{10830538} despite the increased density. The entire $9$--$15$ exploration completes in $28$~minutes, demonstrating that diverse signal--ground assignments can be evaluated rapidly to discover configurations with improved crosstalk performance.}

\greenrev{Such non-intuitive, high-density layouts are typically out of reach for FEM-bound co-simulation workflows~\cite{10830538,10198347,10403538,10816184}, since exhaustive Pareto search over millions of configurations is only tractable at surrogate cost.}

\rev{
\subsubsection{Optimization Beyond Exhaustive Search}
For larger arrays, exhaustive placement enumeration becomes intractable even after $D_4$ symmetry reduction. We therefore use a surrogate-driven NSGA-II search, where candidate placements are evaluated by TSV-PhGNN at millisecond cost; the optimizer can be replaced by other black-box methods such as GA, simulated annealing, or Bayesian optimization.

The search is evaluated on $N{\times}N$ arrays with $N\in\{7,9,11\}$ using $10{,}000$ surrogate evaluations over three seeds. The placement space $|\mathcal{S}|={}_{N^2}C_{n_{\mathrm{sig}}}$, with $n_{\mathrm{sig}}=\mathrm{round}(0.48N^2)$, grows from $\sim 6.5{\times}10^{13}$ for $7{\times}7$ to $\sim 4.4{\times}10^{35}$ for $11{\times}11$. As shown in Table~\ref{tab:r3_4_optimization}, NSGA-II consistently outperforms random search, achieving hypervolume gains of $1.79\times$, $3.15\times$, and $1.76\times$ for $N=7,9,11$, respectively. The resulting Pareto trends are also supported by Kendall's $\tau\ge 0.78$ across the relevant ranking metrics in Table~\ref{tab:r3_2_suite}.
}

\begin{table}[t]
\centering
\caption{\rev{NSGA-II versus random-search hypervolume at $10{,}000$ surrogate evaluations (3 seeds). Search space $|\mathcal{S}|={}_{N^2}C_{n_{\text{sig}}}$, $n_{\text{sig}}=\mathrm{round}(0.48\,N^2)$; objectives: $-\overline{|S_{21}|}$, $\max|S_{11}|$, $\max|\text{NEXT}|$, $\max|\text{FEXT}|$ in dB.}}
\label{tab:r3_4_optimization}
\footnotesize
\setlength{\tabcolsep}{3pt}
\renewcommand{\arraystretch}{1.15}
\begin{tabular}{@{}c l l l c@{}}
\toprule
$N$ &
\makecell[l]{Search\\space $|\mathcal{S}|$} &
\makecell[l]{NSGA-II HV\\mean\ [min,\,max]} &
\makecell[l]{Random HV\\mean\ [min,\,max]} &
Ratio \\
\midrule
$7\!\times\!7$
  & \makecell[l]{${}_{49}C_{24}$\\$\approx 6.5{\times}10^{13}$}
  & \makecell[l]{0.311\\{\scriptsize[0.288,\,0.345]}}
  & \makecell[l]{0.173\\{\scriptsize[0.076,\,0.255]}}
  & $1.79\times$ \\[2pt]
$9\!\times\!9$
  & \makecell[l]{${}_{81}C_{39}$\\$\approx 8.1{\times}10^{22}$}
  & \makecell[l]{0.431\\{\scriptsize[0.287,\,0.541]}}
  & \makecell[l]{0.137\\{\scriptsize[0.109,\,0.187]}}
  & $3.15\times$ \\[2pt]
$11\!\times\!11$
  & \makecell[l]{${}_{121}C_{58}$\\$\approx 4.4{\times}10^{35}$}
  & \makecell[l]{0.797\\{\scriptsize[0.718,\,0.862]}}
  & \makecell[l]{0.453\\{\scriptsize[0.106,\,0.685]}}
  & $1.76\times$ \\
\bottomrule
\end{tabular}
\end{table}

% \rev{\subsubsection{Geometrical Pareto Sweep}}

\greenrev{The proposed framework also supports TSV geometrical optimization beyond spatial placement. To demonstrate this capability, a multi-objective Pareto optimization is performed over the TSV radius, pitch, height, and oxide thickness ranges in Table~\ref{tab:tsv_parameters_sweep}. For each sampled design, the surrogate evaluates electrical metrics, including $\max |S_{11}|$, average insertion loss $\overline{S_{21}}$, and worst-case NEXT/FEXT, while the ETC model computes the effective thermal conductivities $(k_x,k_y,k_z)$.}

\begin{table}[b]
\centering
\vspace{-10pt}
\caption{Geometrical Parameters of TSV Array Sweep}
\label{tab:tsv_parameters_sweep}
\begin{tabular}{ccc}
\hline\hline
\textbf{Parameters} & \textbf{Symbol} & \textbf{Value} \\
\hline
TSV radius & $r_{\mathrm{TSV}}$ & $2\rightarrow6~\mu\mathrm{m}$ \\
TSV height & $h_{\mathrm{TSV}}$ & $60\rightarrow100~\mu\mathrm{m}$ \\
Pitch between adjacent TSVs & $p$ & $20\rightarrow60~\mu\mathrm{m}$ \\
Thickness of SiO$_2$ layer & $t_{\mathrm{ox}}$ & $0.5\rightarrow3~\mu\mathrm{m}$ \\
\hline\hline
\vspace{-10pt}
\end{tabular}
\end{table}
\greenrev{Constructing the Pareto frontier across these competing objectives enables tradeoff analysis between signal integrity (predicted by the ML model) and thermal performance (computed via the ETC-based framework of Section~\ref{electrothermal_model}). The Pareto extraction yields four representative optimal configurations, each optimizing a distinct objective (crosstalk, thermal conductivity, insertion loss, or reflection); the corresponding parameters and metrics are summarized in Table~\ref{tab:pareto_best_designs}.}

As shown in Table~\ref{tab:pareto_best_designs}, the optimal configurations expose clear tradeoffs: crosstalk-optimal designs reach $-50.7$~dB coupling but not maximum thermal conductivity, thermal-optimal designs push $k_z$ to $149$~W/mK at the cost of crosstalk, the insertion-loss-optimal design reaches $\overline{S_{21}}=-0.066$~dB, and the reflection-optimal design improves matching to $\max S_{11}=-45.75$~dB without simultaneously minimizing coupling.

\begin{table}[ht]
\centering
\setlength{\tabcolsep}{6pt}
\caption{Pareto-Optimal TSV Designs at 15~GHz}
\label{tab:pareto_best_designs}
\begin{tabular}{lcccc}
\hline\hline
\textbf{Parameter} 
& \shortstack{\textbf{Best}\\\textbf{Crosstalk}} 
& \shortstack{\textbf{Best}\\\textbf{Thermal}} 
& \shortstack{\textbf{Best}\\\textbf{Insertion}} 
& \shortstack{\textbf{Best}\\\textbf{Reflection}} \\
\hline

$r_{\mathrm{TSV}}$ ($\mu$m) 
& 3.56 & 2.63 & 2.03 & 2.69 \\

$p$ ($\mu$m) 
& 20.15 & 53.43 & 29.66 & 20.09 \\

$h_{\mathrm{TSV}}$ ($\mu$m) 
& 62.20 & 60.04 & 60.74 & 60.17 \\

$t_{\mathrm{ox}}$ ($\mu$m) 
& 2.72 & 0.90 & 2.98 & 2.98 \\

$\overline{S_{21}}$ (dB) 
& -0.083 & -0.171 & -0.066 & -0.071 \\

$\max S_{11}$ (dB) 
& -35.29 & -35.74 & -35.30 & -45.75 \\

Worst Xtalk (dB) 
& -50.71 & -40.10 & -40.62 & -42.12 \\

$k_z$ (W/mK) 
& 142.35 & 149.02 & 140.72 & 133.81 \\

\hline\hline
\end{tabular}
\end{table}

Unlike iterative full-wave co-simulation, the surrogate-based framework enables efficient exploration of the multi-dimensional design space and lets designers select TSV configurations by application-specific priority (crosstalk, matching, insertion loss, or thermal conductivity), turning TSV design from trial-and-error into a systematic data-driven optimization.

\vspace{-5pt}
\subsection{Scalability and Computational Considerations}

% Computational scalability of the proposed graph-based ML framework is discussed in this section. 
% For an $N \times N$ TSV grid, the total number of TSVs scales as $N^2$. 
% %If the grid dimension is doubled, i.e., $N \times N \rightarrow 2N \times 2N$, the number of TSVs increases quadratically from $N^2$ to $4N^2$.
% %
% Since the proposed graph representation models coupling dependencies between all TSV pairs,
% the graph is treated as fully connected. 
% Thus, the number of edges for $V = N^2$ TSVs scales as
% %
% \begin{equation}
% E = V(V-1) = \mathcal{O}(N^4).
% \end{equation}
% %
% %where $V = N^2$ is the number of TSVs. 
% %Therefore, the total number of edges scales approximately as $\mathcal{O}(N^4)$.

% This quadratic growth in the number of vertices and quartic growth in edge interactions 
% directly impacts memory consumption and computational complexity. 
% Because the inference stage is primarily GPU-accelerated, the framework becomes bounded by available VRAM for large-scale arrays. 
% Although switching to CPU-based inference allows handling larger graphs, this comes at the expense of increased runtime and reduced efficiency.

% Therefore, a practical tradeoff exists between array scalability, computational time, 
% and hardware memory constraints. This tradeoff should be considered when extending 
% the framework to wafer-scale or ultra-large interposer designs. Future work will explore sparse graph representations and locality-aware edge pruning to reduce memory complexity while preserving physical fidelity.

For an $N \times N$ TSV grid the fully connected graph has $V = N^2$ nodes and $E = V(V-1) = \mathcal{O}(N^4)$ edges, so peak GPU VRAM dominates the inference budget for large arrays.

\rev{
\subsubsection{Inference Scalability}
Inference scalability is evaluated on a $48$~GB NVIDIA RTX~A6000 in terms of peak GPU memory, throughput, and maximum practical array size. Since memory is dominated by the edge count $N^2(N^2-1)$, the production loader batches designs by a fixed total edge budget rather than by a fixed number of graphs. This keeps peak memory controlled as array size increases.

As shown in Table~\ref{tab:r1_5_scalability}, the edge-budget loader keeps peak GPU memory below $4$~GB for fully populated arrays up to $30{\times}30$, while the fixed-batch loader exhausts the same GPU at $N\ge25$. Although the edge-budget loader slightly reduces throughput at moderate sizes, its memory advantage becomes essential for large arrays. Extrapolating the single-design memory trend gives a practical inference ceiling of approximately $55{\times}55$ TSVs, corresponding to over $3{,}000$ TSVs and about $9.1$~M directed edges on the reference GPU.
}

\begin{table}[t]
\centering
\caption{\rev{Peak GPU memory and throughput versus grid size $N{\times}N$ (NVIDIA RTX~A6000, $48$\,GB). Edge-budget loader (budget $8\,192$ edges) versus fixed-batch loader ($32$ designs/batch); fully populated arrays.}}
\label{tab:r1_5_scalability}
\small
\setlength{\tabcolsep}{3pt}
\begin{adjustbox}{max width=\linewidth}
\begin{tabular}{@{}rrrrrrr@{}}
\toprule
& & & \multicolumn{2}{c}{peak mem (MB)} & \multicolumn{2}{c}{throughput (d/s)} \\
\cmidrule(lr){4-5} \cmidrule(lr){6-7}
$N$ & nodes & edges/design & edge-budget & fixed-batch & edge-budget & fixed-batch \\
\midrule
 3 &   9 &      72 &       47 &        21 & 13\,653 & 6\,752 \\
 5 &  25 &     600 &       46 &        96 &  2\,842 & 6\,438 \\
10 & 100 &  9\,900 &       55 &    1\,419 &     230 &    946 \\
15 & 225 & 50\,400 &      234 &    7\,176 &     113 &    194 \\
20 & 400 & 159\,600 &     719 &   22\,690 &      56 &     61 \\
25 & 625 & 390\,000 &  1\,742 & \textit{OOM} & 25 & \textit{OOM} \\
30 & 900 & 809\,100 &  3\,605 & \textit{OOM} & 12 & \textit{OOM} \\
\bottomrule
\end{tabular}
\end{adjustbox}
\end{table}

\rev{
\subsubsection{Scope of Applicability and Practical Deployment}
The proposed framework targets per-die and per-chiplet TSV analysis within the validated range covered by the analytical pre-training and HFSS fine-tuning data. Its graph formulation supports irregular layouts through variable TSV occupancy, signal/ground assignment, and geometry-dependent edge features, without requiring a fixed grid. The demonstrated scalability is suitable for realistic 3D-IC, HBM-like, and 2.5D-interposer workloads with hundreds to thousands of TSVs. For regimes outside the validated envelope, such as new liner materials, dimensions, or frequency ranges, the workflow can be extended by adding representative training samples. Wafer- or panel-scale systems with $\mathcal{O}(10^5)$--$\mathcal{O}(10^6)$ vias are better handled compositionally at the tile or channel level, with weaker inter-tile coupling treated by package- or interposer-level analysis.
}

Future work will investigate sparse graph representations and locality-aware edge pruning to further reduce memory complexity while preserving accuracy.

% \subsection{Uncertainty Analysis}
% To assess the epistemic uncertainty, we trained an ensemble of models on varying fractions of the dataset. Fig. \ref{fig:learning_curve} shows the learning curve. The narrowing confidence band indicates that the model converges to a stable solution as data volume increases. Furthermore, Fig. \ref{fig:uncertainty_heat} visualizes the variance in S-parameter magnitude predictions across the ensemble.

\section{Conclusion}
\label{conclusion}
% A scalable electro--thermal co-design framework for dense TSV arrays is proposed that integrates physics-based analytical modeling, machine-learning acceleration, and FEM sign-off within a unified flow. The analytical solver computes wideband multi-port S-parameters and effective anisotropic thermal conductivities for arbitrary TSV array sizes and signal--ground distributions. Validation against HFSS using the complex-domain RFE demonstrated typical errors of $5\%$--$10.6\%$ for moderate array sizes, while reducing evaluation time from $\sim$$10^3$ seconds per FEM simulation to sub-second levels.

% A GNN-based surrogate, fine-tuned with a limited HFSS dataset, further improves accuracy, achieving complex-domain RFE below $\sim$2\% and millisecond-level inference, enabling over $10^6\times$ acceleration compared to full-wave simulation for medium-size ($15\times15$) arrays. 

% The framework enables large-scale multi-objective Pareto exploration across geometrical and spatial configurations, evaluating millions of candidates within minutes and revealing explicit tradeoffs between reflection, insertion loss, crosstalk, and thermal conductivity. Benchmark comparisons demonstrate significant improvements in worst-case crosstalk and electro--thermal balancing relative to prior designs.

% Overall, the proposed approach provides a practical pathway for rapid electro--thermal optimization of dense TSV networks while preserving full-wave verification through automated HFSS and Mechanical export. 

A scalable electro--thermal co-design framework for dense TSV arrays is presented, integrating physics-based analytical modeling, GNN-based acceleration, and FEM sign-off. The analytical solver computes wideband multi-port S-parameters and anisotropic thermal conductivities for arbitrary layouts, agreeing with HFSS to $5\%$--$10\%$ RFE while reducing per-evaluation time from $\sim 10^{3}$~s (FEM) to sub-second. A GNN surrogate fine-tuned on a small HFSS dataset further improves accuracy to \rev{mean RFE below $5\%$} with millisecond inference, giving $>10^{6}\times$ acceleration on $15\times15$ arrays. The framework enables large-scale multi-objective Pareto exploration over geometrical and spatial configurations, evaluating millions of candidates within minutes and exposing tradeoffs among reflection, insertion loss, crosstalk, and thermal conductivity; final designs are verified through automated HFSS and Mechanical sign-off.

\section{Acknowledgments}
This work was supported by the Center on Cognitive Multispectral Sensors (CogniSense), one of seven centers in Joint University Microelectronics Program (JUMP) 2.0, a Semiconductor Research Corporation (SRC) program sponsored by the Defense Advanced Research Project Agency (DARPA).\\
The authors used generative artificial intelligence (ChatGPT, OpenAI) solely for language editing and improving writing clarity. All technical content and results were developed and verified by the authors.

\bibliographystyle{IEEEtran}
\bibliography{references}

\begin{IEEEbiography}
[{\includegraphics[
width=1in,
height=1.25in,
clip,
keepaspectratio]{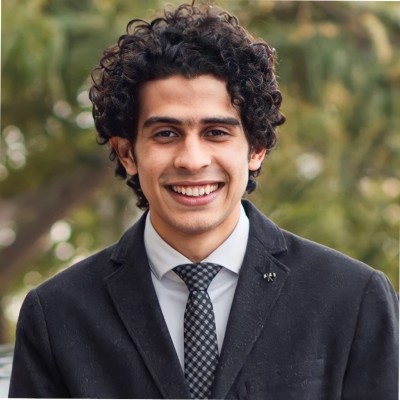}}]{Mohamed Gharib}
(Graduate Student Member, IEEE) received the B.S. degree in Electronics and Electrical Communications Engineering from Cairo University, Cairo, Egypt, in 2021. He is currently pursuing the Ph.D. degree in Electrical and Computer Engineering at the University of Illinois Chicago, where he has been with the HiPerCAS Laboratory since 2023. Prior to his doctoral studies, he gained industry experience in semiconductor physical design. Since 2025, he has also been a Visiting Student with Argonne National Laboratory. His research focuses on physics-based and AI-driven modeling, simulation, and optimization for advanced semiconductor packaging, 2.5D/3D heterogeneous integration, signal/power integrity, and electrothermal reliability.
\end{IEEEbiography}
\vspace{-10pt}
\begin{IEEEbiography}[{\includegraphics[width=1in,height=1.25in,clip,keepaspectratio]{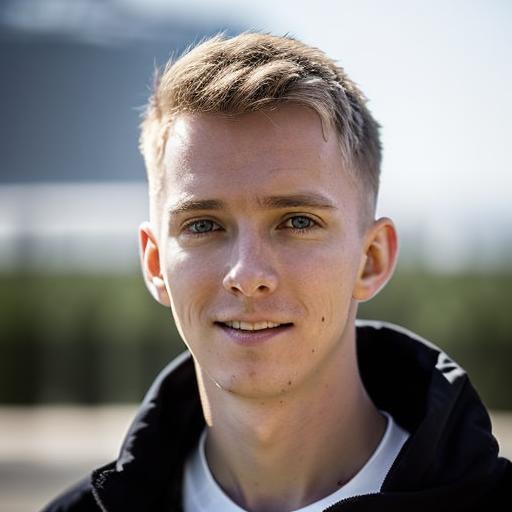}}]{Leonid Popryho}
(Graduate Student Member, IEEE) received the B.Sc. degree in applied mathematics from Igor Sikorsky Kyiv Polytechnic Institute, Kyiv, Ukraine, in 2021, and the dual M.Sc. degree in computer science from Blekinge Institute of Technology, Sweden, and Kyiv Academic University, Ukraine, in 2023. He is currently pursuing the Ph.D. degree in electrical and computer engineering at the University of Illinois Chicago, IL, USA.
Since 2023, he has been a Graduate Research Assistant with the HiPerCAS Laboratory, University of Illinois Chicago, and in 2026 joined the X-ray Science Division, Argonne National Laboratory, Lemont, IL, USA, as a Research Aide. His research interests include machine learning for electronic design automation, graph neural networks, physics-informed neural networks, active learning, and surrogate modeling for device and circuit optimization.
He was a recipient of the DAC Young Fellowship in 2025 and 2026.
\end{IEEEbiography}
\vspace{-11pt}
\begin{IEEEbiography}[{\includegraphics[
width=1in,
height=1.25in,
clip,
keepaspectratio]{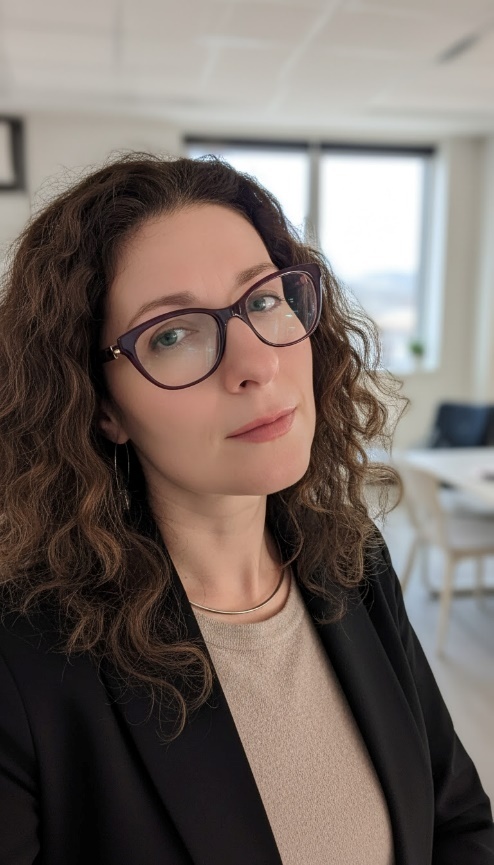}}]{Inna Partin-Vaisband}
(Senior Member, IEEE) received the Ph.D. degree in Electrical and Computer Engineering from the University of Rochester in 2015, and the B.Sc. and M.Sc. degrees from the Technion–Israel Institute of Technology in 2006 and 2009, respectively. She is currently an Associate Professor of Electrical and Computer Engineering at the University of Illinois Chicago. She was with IBM in 2005–2009. Her research focuses on enabling 2.5D/3D high-performance computing platforms through co-design methodologies and automation and power delivery. Dr. Partin-Vaisband serves as an Associate Editor for the \textit{IEEE Transactions on Components, Packaging and Manufacturing Technology} and \textit{IEEE Circuits and System Magazine}. She is a recipient of the 2022 Google Research Scholar Award and the 2023 NSF CAREER Award.
\end{IEEEbiography}
\end{document}